\documentclass[lettersize,journal]{IEEEtran}
\usepackage{amsmath,amsfonts}
\usepackage{algorithmic}
\usepackage{algorithm}
\usepackage[caption=false,font=normalsize,labelfont=sf,textfont=sf]{subfig}
\usepackage{textcomp}
\usepackage{stfloats}
\usepackage{url}
\usepackage{verbatim}
\usepackage{graphicx}
\usepackage{cite}
\usepackage{xcolor}
\usepackage{multirow}
\usepackage{booktabs}
\usepackage{array}
\usepackage{subcaption} 
\begin{document}

\title{Accurate Motion Estimation with Bézier Control Point for Efficient Frame Interpolation}

\author{Shuhao Han, Chenyang Wu, Chun-Le Guo, Zheng-Peng Duan, Zhen Li, Ming-Ming Cheng, Chongyi Li
\thanks{
    This work was supported in part by the Tianjin Natural Science Foundation Project (25ZXRGGX00290, 24JCJQJC00020, 25JCQNJC01390), National Natural Science Foundation of China (62306153, 62225604), the Young Elite Scientists Sponsorship Program by CAST (YESS20240686), the Fundamental Research Funds for the Central Universities (Nankai University, 63243143, 63253223, 63253219, 63261190), and Shenzhen Science and Technology Program (JCYJ20240813114237048). The computational devices were supported by the Super computing Center of Nankai University (NKSC). This work was also supported by OPPO Research Fund. (Corresponding author: Chongyi Li.)}
\thanks{
    Shuhao Han, Chenyang Wu, and Zheng-Peng Duan are with VCIP,
    College of Computer Science, Nankai University, Tianjin 300350,
    China (e-mail: hansh@mail.nankai.edu.cn; chenyangwu@mail.nankai.edu.cn; adamduan0211@mail.nankai.edu.cn) . Shuhao Han and Chenyang Wu contributed equally to this work. 
}
\thanks{
    Chun-Le Guo, Ming-Ming Cheng, and Chongyi Li are with VCIP, College of Computer Science,
    Nankai University, Tianjin 300350, China, and also with the Nankai International Advanced Research Institute (NKIARI), Futian, Shenzhen 518045,
    China (e-mail: guochunle@nankai.edu.cn; cmm@nankai.edu.cn; lichongyi@nankai.edu.cn). 
}
\thanks{
    Zhen Li is with Alibaba Group, Hangzhou 311121, China (e-mail: zhenli1031@gmail.com).
}
}

\maketitle

\begin{abstract}

In frame interpolation tasks, motion ambiguity in the training set causes models to generate blurry intermediate frames.
Moreover, the assumption of uniform motion between frames during inference further leads to inaccuracies in the generated intermediate frames. 
To tackle these challenges, we propose an \textbf{A}ccurate motion estimation algorithm with \textbf{B}ézier \textbf{C}ontrol point, ABC-Inter, for efficient frame \textbf{Inter}polation.
Specifically, ABC-Inter designs an \textbf{A}ccurate \textbf{F}low estimation \textbf{M}odule (AFM) by decoupling two-frame features and mapping to corresponding coordinates to better estimate the optical flow between the two frames.
Furthermore, ABC-Inter eliminates motion ambiguity in the training set by introducing Bézier control points that are computed using the input frames and the intermediate ground-truth (gt) frames.
This allows the model to estimate accurate optical flow between two frames during the training process, thereby solving the blurriness problem in the generated intermediate frames during inference.
Benefiting from the more accurate flow estimation between two frames, we can introduce additional frames and directly use multiple flows to calculate Bézier control points for modeling non-uniform motion without retraining the model. 
Simultaneously, to realize the estimation of non-linear motion using only two frames, we also introduce a new Bézier control point estimation module which achieves better motion estimation between the two frames by performing fine-tuning on the model in the second stage.
Experimental results demonstrate that our ABC-Inter achieves state-of-the-art performance on multiple benchmark datasets and exhibits excellent visual perception.

\end{abstract}

\section{Introduction}

Video frame interpolation (VFI) is a classic task in the field of multimedia processing~\cite{kye2026acevfi,chen2023snis, chen2025fpsmark, chen2025prest, zhang2025hfct}. Its primary goal is to synthesize intermediate frames between consecutive video frames, thereby increasing the video's frame rate. VFI has widespread applications in various downstream tasks, including slow-motion generation~\cite{jiang2018super,xiang2020zooming}, novel viewpoint synthesis~\cite{flynn2016deepstereo, li2021neural}, animation creation~\cite{siyao2021deep}, and video compression~\cite{wu2018video}. Recently, it has shown promising potential in the smart transition (SAT) task of multi-camera systems in smartphones.%

Flow-based methods \cite{jin2023unified,zhang2023extracting} have become mainstream in the field of VFI, which involve estimating the motion between two frames and then warping frames to the intermediate moment. Consequently, the quality of interpolation results heavily relies on the accuracy of flow estimation. %

Based on the warping approach employed for intermediate frame generation, flow-based methods can be categorized into two types: backward warping and forward warping. Backward warping methods \cite{li2023amt,zhang2023extracting} 
utilize a sampling method to backward warp frames to the intermediate moment, which has been proven to be simple and effective.
However, the challenge lies in obtaining high-quality flow that corresponds to the coordinates of the intermediate invisible frame.
Additionally, its alignment with the intermediate frame restricts the model's capability to handle arbitrary frame interpolation.
Even though arbitrary frame interpolation can be achieved via time $t$ injection, the optical flow needs to be recalculated for each specific time $t$, which increases the computational latency.
Forward warping methods~\cite{niklaus2020softmax, jin2023unified} utilize flow aligned with the coordinates of visible frames for splatting. The advantage is free from requiring coordinate transformation for the flow, making the flow estimation more straightforward. Moreover, the estimated flow can be linearly scaled to obtain motion at any given moment, enabling effective interpolation at arbitrary instants. %
Benefiting from this, the forward warping method only requires calculating a dual optical flow between two frames when performing arbitrary frame interpolation, and this dual optical flow can be directly applied to any time $t$ to improve the efficiency of arbitrary frame interpolation.

\begin{figure}[t]
  \includegraphics[width=\linewidth]{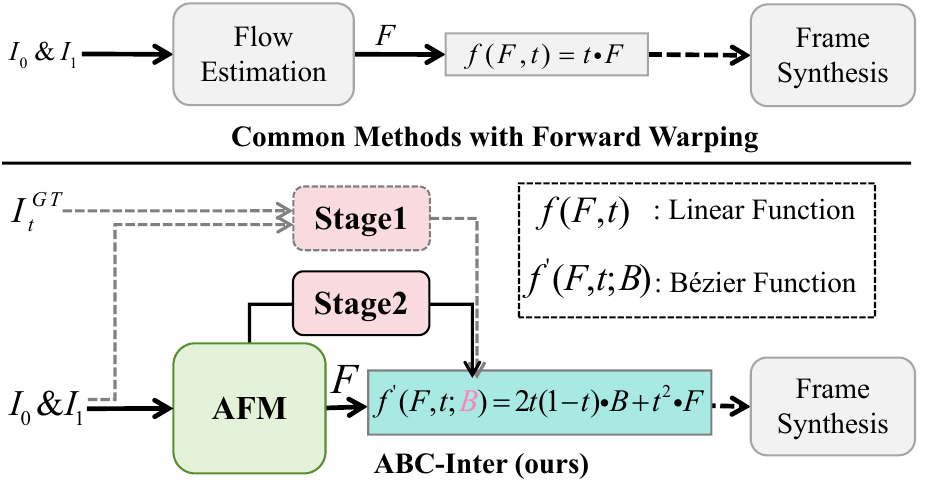}
  \caption{The general frame interpolation framework of common flow-based methods with forward warping and ours. Instead of using a linear function to map the flow to the intermediate frame, our approach redesigns an Accurate Flow estimation Module (AFM) and introduces Bézier control points to change the computation of intermediate frame motion and employs a two-stage strategy.}%
  \label{fig:first}
\end{figure}

In forward warping methods, accurately estimating the motion between visible frames and mapping it onto the intermediate frame for warping is crucial. As shown in Fig.~\ref{fig:first}, previous approaches typically assume uniform linear motion between frames, deriving motion for the intermediate frame by straightforwardly scaling the visible frame's motion based on the time step. However, due to complex non-uniform motion between frames in real-world scenarios, the simplistic assumption of uniform motion might fail in representing various motion patterns, resulting in suboptimal synthesis of intermediate frames.

\begin{figure*}[ht]
  \centering
  \includegraphics[width=\textwidth]{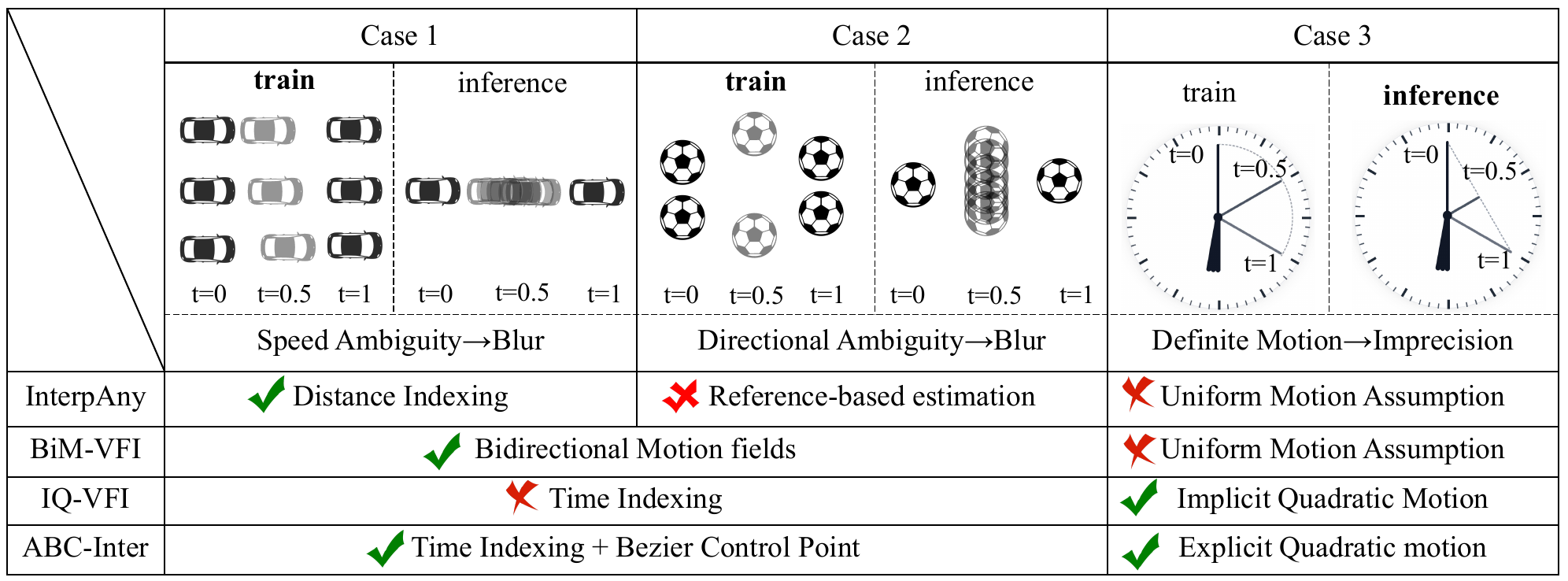}
  \caption{Currently three main issues exist in frame interpolation tasks.
Case 1 and Case 2 indicate that in the training set, objects in the same scene may have ambiguities in terms of speed and motion direction. Such ambiguities make it difficult for the model to estimate the exact motion during inference, thereby leading to the generation of blurry intermediate frames. Case 3, on the other hand, demonstrates a specific curved motion—if only linear motion can be assumed during inference, inaccuracies in the motion of the generated intermediate frames will occur.
InterpAny~\cite{zhong2023clearer} addresses the speed ambiguity by converting to distance indices and alleviates the motion direction ambiguity using a reference-based estimation method. BiM-VFI~\cite{seo2024bim}, meanwhile, uniformly resolves both speed and motion direction ambiguities through bidirectional motion fields. However, the assumption of uniform motion in these two methods during inference makes it difficult for them to solve the problem in Case 3.
IQ-VFI~\cite{hu2024iq} tackles the issue in Case 3 via implicit motion modeling, but it fails to account for motion ambiguities in the training set. In contrast, ABC-Inter resolves motion ambiguities by introducing Bezier Control Points and incorporates a Bezier Control Point estimation module in the second stage to address the uniform motion assumption problem in Case 3.}
  \label{fig:case}
\end{figure*}

This challenge also exists in backward warping-based methods.
As shown in Case 3 of Fig.~\ref{fig:case}, the clock's movement adheres to a specific circular motion. However, assuming uniform motion from $t=0$ to $t=1$ would yield inaccurate inference outcomes.
IQ-VFI~\cite{hu2024iq} can address the issue in Case 3 through implicit quadratic motion modeling.
However, due to the nature of its implicit modeling, it cannot directly incorporate multiple frames to achieve more accurate non-linear modeling. Meanwhile, it still suffers from the problems illustrated in Case 1 and Case 2 of Fig.~\ref{fig:case}.

Case 1 and Case 2 demonstrate that in the training set, the same object in the same scene may exhibit different motion patterns. 
Directly using such data for training will cause the model to be unable to determine which specific motion pattern to model during inference.
This uncertainty leads to the generation of blurry results by the model at inference time.
InterpAny~\cite{zhong2023clearer} eliminates the ambiguity of speed in the training set by converting time indexing into distance indexing. For velocity direction, it adopts a reference-based estimation method to mitigate this issue. BiM-VFI~\cite{seo2024bim}, on the other hand, directly resolves the ambiguities of both speed and velocity direction through bidirectional motion fields.
These methods address challenges in backward warping-based approaches. However, compared to forward warping, backward warping methods are less efficient for arbitrary frame interpolation due to iterative optical flow estimation.
Moreover, by only modeling three-frame dataset positions during training without considering specific motion pattern, these methods rely on linear motion assumptions at inference, failing to handle complex scenarios like Case 3 in Fig~\ref{fig:case}.

To tackle these challenges, we propose an \textbf{A}ccurate motion estimation algorithm with \textbf{B}ézier \textbf{C}ontrol point for efficient frame Interpolation, abbreviated as ABC-Inter. 

As a frame interpolation method based on forward warping, we first design a simple yet effective \textbf{A}ccurate \textbf{F}low estimation \textbf{M}odule (AFM) to better estimate bidirectional flow between input frames. This module employs a coarse-to-fine strategy to accurately estimate bidirectional flow between input frames like UPR-Net~\cite{jiang2018super}. Different from UPR-Net~\cite{jiang2018super}, which leverages coarse optical flow to warp input features to the intermediate frame for flow refinement, ABC-Inter decouples the estimation of bidirectional optical flow between two frames. Specifically, it uses the coarse optical flow to warp the feature of one frame to the coordinate space of the other input frame for update, thereby ensuring the consistency of coordinates in bidirectional optical flow estimation.

Second, to address the ambiguities of speed and motion direction in the training set, we introduce Bézier control points by modeling the motion patterns among three frames to eliminate motion ambiguity in the training set.
Compared with InterpAny~\cite{zhong2023clearer} and BiM-VFI~\cite{seo2024bim} which converts time indexing to distance indexing or bidirectional motion fields to model the positions of three frames in the training set for ambiguity elimination, ABC-Inter retains concept of time indexing. 
Moreover, during arbitrary frame interpolation inference, ABC-Inter only needs to calculate the bidirectional optical flow between two frames once. 
In contrast, InterpAny~\cite{zhong2023clearer} and BiM-VFI~\cite{seo2024bim} require separate optical flow estimation for arbitrary interpolation during inference, which is far more time-consuming.

Beyond efficiency advantages, ABC-Inter also outperforms InterpAny and BiM-VFI in motion modeling capability: while InterpAny and BiM-VFI can only assume uniform linear motion during inference, ABC-Inter enables non-uniform motion modeling at inference time by introducing additional frames to estimate multiple flows for Bézier control point calculation.
Furthermore, benefiting from its more accurate optical flow estimation, ABC-Inter achieves more significant performance gains when using multiple frames for interpolation compared to other forward warping methods such as UPR-Net~\cite{jin2023unified}.

Additionally, to achieve better motion estimation when only two frames are provided as input, we introduce an extra Bézier control point estimation module. This module realizes the prediction of Bézier control points by fine-tuning the previously trained model in the second stage. Through this two-stage training strategy, our final model achieves excellent results across multiple benchmark datasets.

In summary, our key contributions are as follows:
1) designing an optical flow estimation module to estimate the bidirectional optical flow between two frames;
2) introducing Bézier control points to address motion ambiguity in the training set;
3) enabling the modeling of complex motions by introducing multiple frames without the need for model retraining;
4) introducing a Bézier control point estimation module and fine-tuning the model in the second stage to achieve improved performance.

\section{Related Work}
\subsection{Video Frame Interpolation}
A large number of video frame interpolation methods based on deep learning have appeared. 
VFI methods can be roughly categorized into four categories: kernel-based~\cite{cheng2020video, cheng2021multiple, lee2020adacof, niklaus2017video}, hallucination based~\cite{choi2020channel, kalluri2023flavr, kim2020fisr,shi2022video, gui2020featureflow, xiang2020zooming, shen2020blurry}, flow-based~\cite{kong2022ifrnet, niklaus2020softmax, li2023amt, jin2023unified, hu2024iq, wu2024perception, liu2024sparse, liu2023jnmr, liu2023ttvfi, kong2023dynamic, chen2026ifevfi} and diffusion-based~\cite{danier2024ldmvfi, jain2024video, hai2025hierarchical, peng2026ldfvfi, lyu2025tlb, guo2025controllable} methods. 

\noindent
\textbf{Kernel- and Hallucination-based methods.} Kernel-based methods mainly capture motion through dynamic kernel weights~\cite{niklaus2017video, niklaus2017video2, peleg2019net} or offsets~\cite{cheng2020video, cheng2021multiple, ding2021cdfi, lee2020adacof}. Hallucination-based methods use an off-the-shelf architecture~\cite{dai2017deformable, hu2018squeeze, shi2016real, tran2015learning} to generate interpolated frames directly from the features of the input frames. Some methods, like FeFlow~\cite{gui2020featureflow} and Zooming Slow-Mo~\cite{xiang2020zooming}, use deformed convolution~\cite{dai2017deformable} to interpolate intermediate frames in the feature domain. However, kernel-based methods are usually efficient but have limited ability to handle large and complex motion due to the restricted receptive field of local kernels. Hallucination-based methods can synthesize visually plausible frames, but they may lack explicit motion modeling and suffer from structural inconsistency in challenging motion scenarios.  

\noindent
\textbf{Flow-based methods.} Flow-based methods have become the mainstream of VFI research due to the excellent motion modeling ability of the optical flow. In recent studies, the models usually adopt two standard schemes of backward warping~\cite{bao2019memc, huang2022real, kong2022ifrnet} or forward warping~\cite{niklaus2018context, niklaus2020softmax, niklaus2021revisiting} to generate interpolated frames.

\textit{Backward warping-based methods} are considered to be a competitive method, but the pixel overlap problem affects the quality of interpolated frames.
Benefiting from the design of bi-directional correlation volumes for all pairs of pixels, AMT~\cite{li2023amt} reduces the difficulty of modeling large motions and dealing with occluded regions during frame interpolation. 
EMA-VFI~\cite{zhang2023extracting} explicitly separates motion and appearance information by creating a hybrid pipeline, which allows efficient extraction of motion and appearance features without losing fine-grained information.
While backward warping is effective, it remains a challenge to estimate flow due to the need to align coordinates to the intermediate frames. 

\textit{Forward warping-based methods} are commonly viewed as a promising approach for their ability to explicitly model motion, although they can introduce holes.
SoftSplat~\cite{niklaus2020softmax} improves the quality of images produced by forward warping interpolation through an innovative form of splatting. 
M2M-VFI~\cite{hu2022many} estimates multiple bi-directional flows, which establishes a many-to-many splatting scheme with robustness to artifacts like holes. 
UPR-Net~\cite{jin2023unified} uses lightweight recurrent modules for both bi-directional flow estimation and intermediate frame synthesis. We follow the basic framework in UPR-Net~\cite{jin2023unified} but modify it to better adapt to our Bézier control point estimation for frame interpolation.

\noindent
\textbf{Diffusion-based methods.} 
With the advancement of generative model~\cite{ho2020denoising}, recent works have applied diffusion-related techniques to video frame interpolation (VFI).
TLB-VFI~\cite{lyu2025tlb} leverages a temporally aware latent Brownian bridge diffusion mechanism for precise intermediate frame generation in the latent space, while Eden~\cite{zhang2025eden} enhances spatiotemporal feature fusion and motion constraints to mitigate misalignment and artifacts in large-motion scenarios.
Although diffusion-based methods can improve visual coherence and realism in challenging scenes, their multi-step denoising process makes them much slower than optical flow-based methods, limiting their practicality for real-time applications.

\subsection{Variable Velocity Motion Modeling}
In video processing tasks, complex motion patterns can make it difficult for linear optical flow to perform optimally in areas such as video stabilization~\cite{liu2021hybrid}, video super-resolution~\cite{wang2019edvr}, and motion estimation~\cite{bao2019memc}. Specifically, in the field of VFI, the limitations of linear optical flow similarly affect the model's ability to accurately estimate variable velocity motion. 
\cite{xu2019quadratic} uses adjacent multiple frames to compute a quadratic curve model to fit the curved motion of an object, but it has some limitations due to the need for multiple frames involved in the computation. \cite{zhong2023clearer} improves the clarity of interpolated frames by transforming the time index into the distance index to solve the velocity ambiguity problem during training. 
~\cite{seo2024bim} adds an offset angle to ~\cite{zhong2023clearer} to resolve directional ambiguity.
\cite{hu2024iq} adopts implicit quadratic motion estimation for video frame interpolation.
Inspired by Bézier curves and considering the advantages of forward warping for explicitly modeling motion,  we introduce Bézier control points to transform the linear flow into curved for modeling the variable velocity motion better in our ABC-Inter.
Compared with existing non-linear motion modeling methods, ABC-Inter introduces Bézier control points to eliminate motion ambiguity in the training set and ensure the accuracy of optical flow between two frames. 
On this basis, better results can be achieved in non-linear motion modeling whether multiple frames are used to estimate Bézier control points or a network is employed to estimate Bézier control points.

\section{Methodology}
\begin{figure*}[ht]
  \centering
  \includegraphics[width=\textwidth]{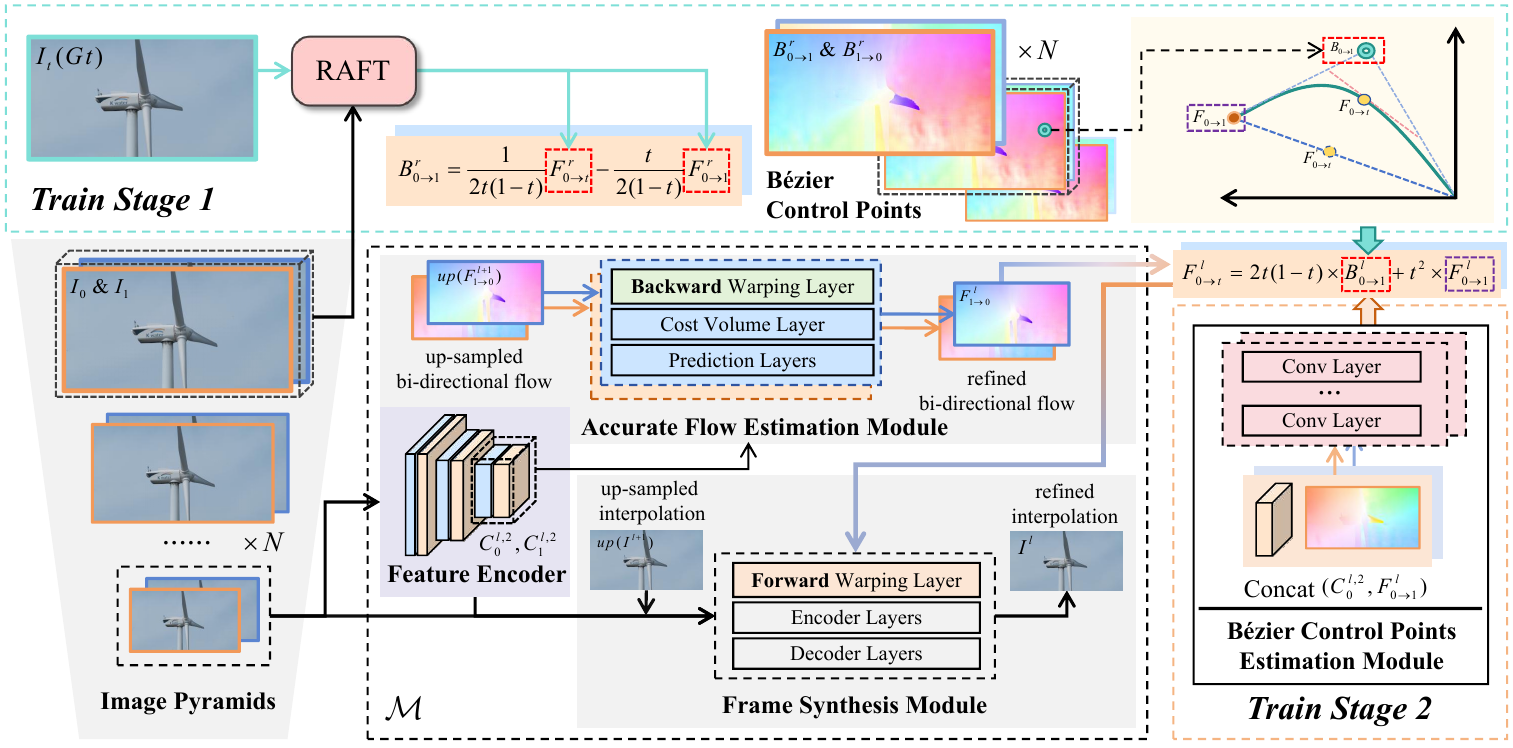}
  \caption{Architecture overview of the proposed ABC-Inter. Our model adopts a two-stage training process. In training stage one, we first use an off-the-shelf  optical flow network RAFT~\cite{teed2020raft} and intermediate frame $I_{t}$ to estimate two optical flows $F_{0 \rightarrow t},F_{0 \rightarrow 1}$, and then calculate the corresponding control points of Bezier and apply them to generate intermediate optical flows. In training stage two, an additional module is introduced to estimate control points. Using the model parameters trained in the first stage as initialization, an optimization strategy is employed to retrain our model.}
  \label{fig:pipeline}
\end{figure*}

\subsection{Overview}

Our ABC-Inter adopts the pyramid recurrent network used in UPR-Net~\cite{jin2023unified}. The overall structure of our method is shown in Fig.~\ref{fig:pipeline}. 
First, we downsample the input images $I_0, I_1$ to build an image pyramid. Second, for each layer of the image $I_0^l,I_1^l$, there is a complete model $\mathcal M$ to generate the intermediate frames $I_t^l$:
\begin{equation}
I_t^l, F_{0 \to 1}^l, F_{1 \to 0}^l = {\mathcal M}(I_0^l, I_1^l, I_t^{l+1}, F_{0 \to 1}^{l+1}, F_{1 \to 0}^{l+1}),
\label{eq:M}
\end{equation} 
where the generation of intermediate frame $I_t^l$ and flows $F_{0 \to 1}^l, F_{1 \to 0}^l$ at each layer utilizes $I_0^l, I_1^l$ and the bilinear upsampling of $I_t^{l+1}$, $F_{0 \to 1}^{l+1}$, $F_{1 \to 0}^{l+1}$ from the previous layer. In particular, $F_{0 \to 1}^{l+1}, F_{1 \to 0}^{l+1}$ is initialized to zero at top level, while $I_t^{l+1}$ is set to the weighted sum of two warped frames with respect to time $t$. Then, the model $\mathcal M$ is iteratively used $N$ times until the final frame $I_t^0$ is generated.

At each layer of the model, $I_0^l,I_1^l$ are fed into a feature encoder to extract three-layer features $\{C_0^{l,0},C_1^{l,0}\}$, $\{C_0^{l,1},C_1^{l,1}\}$, $\{C_0^{l,2},C_1^{l,2}\}$. Then, using the last layer feature $\{C_0^{l,2},C_1^{l,2}\}$ and the flows $F_{0 \to 1}^{l+1}, F_{1 \to 0}^{l+1}$ from the previous layer, the flows $F_{0 \to 1}^l, F_{1 \to 0}^l$ are further updated. Next, the Bézier flow control points are utilized to map them to the intermediate flows $F_{0 \to t}^l, F_{1 \to t}^l$ . Subsequently, $F_{0 \to t}^l, F_{1 \to t}^l$ is employed to warp $I_0^l,I_1^l$  and $\{C_0^{l,0},C_1^{l,0}\}$, $\{C_0^{l,1},C_1^{l,1}\}$, $\{C_0^{l,2},C_1^{l,2}\}$ to a specific time instance using forward warping. The warped results are then concatenated with the ${I_t^{l+1}}$ from the previous layer and passed through an Encoder-Decoder network to obtain the output $M_0^l,M_1^l,R^l$. We then use the following formula to calculate $I_t^l$:
\begin{equation}
    I_t^l = \frac{{(1 - t) \cdot M_0^l \odot I_{0 \to t}^l+ t \cdot M_1^l \odot I_{1 \to t}^l}}{{(1 - t) \cdot M_0^l + t \cdot M_1^l}} + {R^l},
\label{eq:It}
\end{equation}
where $I_{0 \to t}^l = {\mathcal W}({I_0^l},{F_{0 \to t}^l}),I_{1 \to t}^l = {\mathcal W}({I_1^l},{F_{1 \to t}^l})$.   $\mathcal W$ denotes the operation of forward warping. $\odot$ denotes element-wise multiplication. $M_0^l,M_1^l$ are the estimated occlusion masks of $I_{0 \to t}^l$ and $I_{1 \to t}^l$. ${R^l}$ is the estimated residual image. 

To enhance the precision of motion pattern estimation, our ABC-Inter introduces three elaborate designs: 1) designing a simple yet effective flow estimation module; 2) introducing Bézier control points to address motion ambiguity in the training set; 3) using two methods to compute Bézier control points for modeling non-uniform motion during inference.

\subsection{Accurate Flow Estimation Module}

\begin{figure*}[t]
  \centering
  \includegraphics[width=\linewidth, height=5cm]{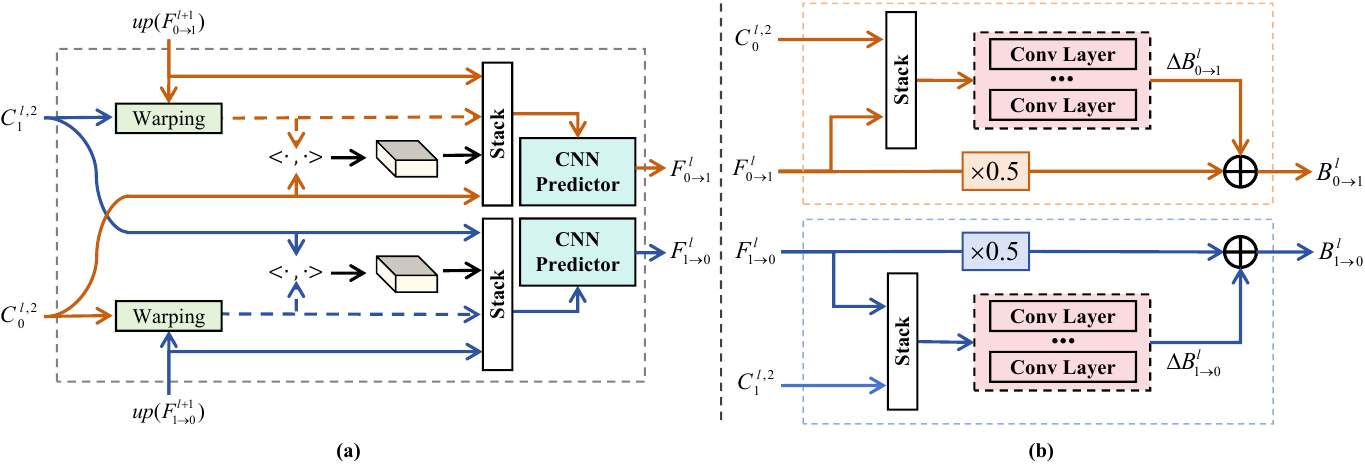}
  \caption{(a) Accurate Flow estimation Module (AFM). The ``Warping'' in the module denotes backward warping. The two ``CNN predictor'' share parameters. The symbol $\langle \cdot, \cdot \rangle$ denotes the local correlation operation for constructing the correlation volume. (b) Bézier Control point estimation Module (BCM). The Conv Layer also  share parameters. }%
  \label{fig:flow}
\end{figure*}

To better estimate the motion between two frames, we design a simple and effective \textbf{A}ccurate \textbf{F}low estimation \textbf{M}odule (\textbf{AFM}). The specific structure is shown in Fig.~\ref{fig:flow} (a). Firstly, the updated flow $F_{0 \to 1}^{l+1}, F_{1 \to 0}^{l+1}$  from the previous layer is upsampled using bilinear interpolation.  Then, backward warping is applied to features $C_1^{l,2},C_0^{l,2}$ extracted from the feature encoder separately to align the features. Taking $C_1^{l,2}$ as an example, it is warped to time $0$ using flow $up(F_{0 \to 1}^{l+1})$ to obtain $C_{0\_{warp}}^{l,2}$. Subsequently, a partial correlation volume ~\cite{sun2018pwc} is constructed using $C_{0}^{l,2}$ and $C_{0\_warp}^{l,2}$:
\begin{align}
    CV_0^{l}(x_0, x_1) = \frac{1}{d}C_{0}^{l,2}(x_0)^{T}C_{0\_warp}^{l,2}(x_1),
\label{eq:cv}
\end{align}
where $||x_1 - x_0||_{\infty}\leq r$, $x_0 {\small =} (u_0,v_0), x_1 {\small =} (u_1,v_1)$ are the pixel coordinates in $C_{0}^{l,2}, C_{0\_warp}^{l,2}$, respectively.  Here, \(r\) is the radius of the local search window used for constructing the partial correlation volume. Since $C_{0}^{l,2}$, $C_{0\_warp}^{l,2} \in \mathbb{R}^{d {\small \times} \frac{H^l}{4} {\small \times} \frac{W^l}{4}}$, the resulting $CV_0^{l}$ has dimensions of $(2r+1)^2 {\small \times} \frac{H^l}{4} {\small \times} \frac{W^l}{4}$, where $(H^l,W^l)$ are the shape of the $l$-th layer image.

Subsequently, the calculated $CV_0^{l}$ is concatenated with $up(F_{0 \to 1}^{l+1})$, $C_0^{l,2}$ and  $C_{0 \_ warp}^{l,2}$ warped from $C_1^{l,2}$. These are then collectively fed into a CNN prediction module to obtain an updated flow $F_{0 \to 1}^{l}$.  The flow $F_{0 \to 1}^{l}$ is at 1/4 resolution of the input frames. Therefore, we upsample $F_{0 \to 1}^{l}$ using bilinear interpolation to the original image size. Simultaneously, $F_{1\to 0}^{l}$ is updated using the same method.

\begin{figure}[t]
    \centering
    \includegraphics[width=\linewidth]{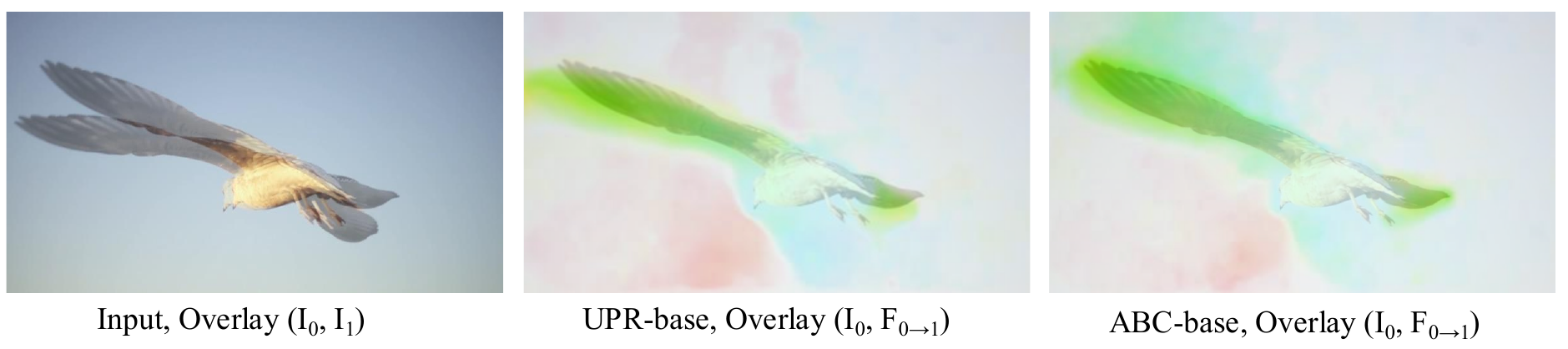}
    \caption{Flow visualization results of UPR-Net and ABC-Inter. It can be observed that the flow $F_{0\rightarrow1}$ computed by ABC-Inter exhibits better coordinate alignment with the corresponding input frames $I_{0}$.}
    \label{fig:afm_comp}
\end{figure}
Compared to UPR-Net~\cite{jin2023unified}, we do not forward-warp two features $C_{0}^{l,2},C_{1}^{l,2}$ to the middle frame. Instead, we use backward warping to align the features $C_{1}^{l,2}$ and $C_{0}^{l,2}$ to the coordinates at time $0$ and $1$, and then update the flow $F_{0\to 1}^{l}$ and $F_{1\to 0}^{l}$ separately. By aligning features through warping at corresponding times and estimating bidirectional flow separately, we can make the estimation of bidirectional flow more accurate.
As shown in Fig~\ref{fig:afm_comp}, with the improved Accurate Flow Estimation Module (AFM), optical flow estimated by ABC-Inter demonstrates better coordinate consistency with the corresponding input frames compared to that of UPR-Net.

\subsection{Bézier Flow Control Point}

Frame interpolation involves generating intermediate frames that are not visible in the original sequence. Therefore, it is necessary to map the motion $F_{0 \rightarrow 1},F_{1 \rightarrow 0}$ between two frames to the intermediate frame.
If backward warping is used,  the flow $F_{t \rightarrow 1},F_{t \rightarrow 0}$ would be employed to sample from the input frames. However, due to the necessity of aligning the flow $F_{t \rightarrow 1},F_{t \rightarrow 0}$ with the coordinates of the intermediate frame, this makes the estimation process quite challenging. If using forward warping, since the coordinates of the motion are aligned with the known frame, assuming that the motion between frames $I_{0}$ and $I_{1}$ is uniform, the flow $F_{0 \rightarrow t},F_{1 \rightarrow t}$ can be simply scaled from the flow $F_{0 \rightarrow 1},F_{1 \rightarrow 0}$ at time $t$. Most current methods~\cite{jin2023unified, niklaus2020softmax, hu2022many} using forward warping achieve this in this way. However, the motion in real-world scenarios is diverse. When encountering large motion or long time intervals between two frames, assuming it to be linear motion can lead to substantial errors in the results~\cite{zhong2023clearer}. To address the assumption of constant velocity motion in forward warping and enhance the estimation of $F_{0 \rightarrow t},F_{1 \rightarrow t}$, we propose to use control points from a second-order Bézier curve to introduce variations in the motion between two frames.

The second-order Bézier curve is a widely used mathematical construct in computer graphics and curve modeling. It is defined by three points, including two endpoints $P_{0}, P_{1}$ and one control point $P_{B}$. The curve is smoothly interpolated between the end points $P_{0}, P_{1}$, and the position of the control point $P_{B}$ affects the curvature and direction of the curve. The formula for the second-order Bezier curve is as follows:
\begin{equation}
B(t) = (1-t)^2 \cdot P_0 + 2 \cdot (1-t) \cdot t \cdot P_B + t^2 \cdot P_1 ,
\label{eq:bezier}
\end{equation} 
where $t$ can be represented as a time parameter that varies between $0$ and $1$, $B(t)$ represents the position of the intermediate point at time $t$.

In order to align the Eqn.\eqref{eq:bezier} with the representation of flow, we subtract $P_0$
from both sides of the equation:

\begin{align}
B(t) - P_0  &= 2 \cdot (1 - t) \cdot t \cdot (P_B - P_0) + t^2 \cdot (P_1 - P_0) \nonumber, \\
F_{0 \to t} &= 2 \cdot (1 - t) \cdot t \cdot B_{0 \to 1} + t^2 \cdot F_{0 \to 1}, 
\label{eq:f0t}
\end{align}
where $B_{0 \to 1} = P_B - P_0$ represents the relative Bézier control points applied in the flow, allowing for better estimation of $F_{0 \to t}$. 

It is worth noting that when $B_{0 \to 1} = 0.5 \cdot F_{0 \to 1}$, the Eqn.\eqref{eq:f0t} degenerates to $F_{0 \to t} = t \cdot F_{0 \to 1}$, indicating uniform linear motion at this time. Similarly, an expression for $F_{1 \to t}$ can be obtained:
\begin{equation}
F_{1 \to t} = 2 \cdot (1 - t) \cdot t \cdot B_{1 \to 0} + (1-t)^2 \cdot F_{1 \to 0},
\label{eq:f1t}
\end{equation} 

By introducing Bézier flow control points, we can better simulate the motion of objects between two frames, leading to the improved estimation of flow $F_{0 \to t},F_{1 \to t}$. 

\noindent
\textbf{Elimination of motion ambiguity in the training set}:
Due to the potential existence of diverse motion patterns with the same input frames, it is an ill-posed problem when estimating the motion of the intermediate frame.
Therefore, even if Bézier control points are introduced, if the model is directly used to estimate both optical flow and control points during training, motion ambiguities in the training set will lead to inaccurate estimation of both optical flow and control points, ultimately resulting in the generation of blurry intermediate frames.
To eliminate such motion ambiguities in the training set, we follow the methods~\cite{zhong2023clearer} and~\cite{seo2024bim}, leveraging the three-frame data (including ground truth) in the training set to compute Bézier control points offline.
Specifically, we employ an off-the-shelf optical flow estimation network RAFT ~\cite{teed2020raft} to estimate the $F_{0 \to 1}^{r} $,$F_{0 \to t}^{r}$ for input frames $I_0,I_1$ and the intermediate frame $I_t^{gt}$. Further, according to Eqn.~\eqref{eq:f0t}, we calculate the Bézier flow control points $B_{0 \to 1}^{r}$ using the following expression:
\begin{equation}
B_{0 \to 1}^{r} = \frac{1}{2t(1-t)}F_{0 \to t}^{r} - \frac{t}{2(1-t)}F_{0 \to 1}^{r},
\label{eq:B01r}
\end{equation}

Subsequently, we apply the same method to calculate $B_{1 \to 0}^{r}$.
For each triplet in the training data, with Bézier flow control points $B_{0 \to 1}^{r}, B_{1 \to 0}^{r}$ derived from the ground truth, estimating the motion $F_{0 \to 1} $,$F_{1 \to 0}$ solely between the input frames through the network can accurately determine the intermediate frame's motion $F_{0 \to t} $,$F_{1 \to t}$. This approach bypasses the modeling of motion from input frames to the intermediate frame and mitigates the uncertainty of motion in the training set, resulting in a more precise estimation of flows $F_{0 \to 1}, F_{1 \to 0}$. Similar to ~\cite{zhong2023clearer} and~\cite{seo2024bim}, we can simply assume uniform motion (i.e., setting $B_{0 \to 1} = 0.5 \cdot F_{0 \to 1}$) in inference to generate intermediate frames in the absence of a reference $B_{0 \to 1}^{r}, B_{1 \to 0}^{r}$, which can still yield clearer results.
Benefiting from the more accurate optical flow oriented to the frame interpolation task generated by introducing Bézier control points, we can not only produce much clearer frame interpolation results but also achieve better non-linear motion modeling.

\subsection{Modeling Non-uniform Motion during Inference}
InterpAny and BiM-VFI eliminate motion ambiguities in the training set. However, since their distance indexing and bidirectional motion fields are both implicitly injected into the network to achieve arbitrary frame interpolation, they can only assume uniform linear motion during inference.
In contrast, ABC-Inter explicitly models motion patterns in the training set using Bézier control points to resolve motion ambiguity. This enables non-uniform motion modeling by estimating Bézier control points for arbitrary frame interpolation during inference.
Specifically, we employ two approaches to estimate the Bézier control points.

\noindent
\textbf{Introducing additional frames without requiring retraining.}
Benefiting from the design of the AFM module and the elimination of motion ambiguities in the training set by introducing Bézier control points, ABC-Inter achieves more accurate optical flow estimation for frame interpolation tasks. 
Therefore, we can introduce additional frames during inference to estimate multiple optical flows, thereby directly calculating the Bézier control points. 
Specifically, on the basis of frames $I_0$ and $I_1$, we introduce frame $I_{-1}$, then use the AFM module to calculate the optical flows $F_{0\rightarrow1}$ and $F_{0\rightarrow-1}$ respectively. After that, we calculate $B_{0\rightarrow1}$ using  Eqn.~\eqref{eq:B01r}, and obtain $B_{1\rightarrow0}$ in the same way. ABC-Inter achieves a significant performance improvement through the adoption of multi-frame inference on multi-frame interpolation benchmark.

\noindent
\textbf{Designing a Bézier Control point estimation Module (BCM) to finetune ABC-Inter.}
In addition to using extra frames, we can also design a Bézier Control point estimation Module (BCM) to estimate $B_{0 \to 1},B_{1 \to 0}$, thereby improving the modeling of  complex motion between two frames. The specific structure is shown in Fig.~\ref{fig:flow} (b). Specifically, we initialize the feature extractor, flow estimation module, and frame generation module with the parameters trained in the \textbf{first stage} and add the BCM to the model for retraining in the \textbf{second stage}. 
Regarding the specific design of the BCM, we  initialize control point  $B_{0 \to 1}, B_{1 \to 0}$ to $0.5 \cdot F_{0 \to 1} ,0.5 \cdot F_{1 \to 0}$. Then, the convolutional  network is used to estimate the residual of $\Delta B_{0 \to 1}, \Delta B_{1 \to 0}$. %
We use the output $F_{0 \to 1} ,F_{1 \to 0}$ from the flow estimation module, and the last layer feature $C_0, C_1$ from the feature encoder as input. By passing through four convolutional layers, we obtain the required residual of Bézier flow control points $\Delta B_{0 \to 1}, \Delta B_{1 \to 0}$. The equation is as follows:
\begin{equation}
\Delta B_{i \to 1-i} = \mathcal{C}onvs([C_i,F_{i \to 1-i}]), i\in \{0,1\},
\label{eq:B}
\end{equation}
Therefore, the final expression of the BCM  is as follows: 
\begin{equation}
    B_{i \to 1-i} = 0.5 \cdot F_{i \to 1-i} + \Delta B_{i \to 1-i},
\end{equation}
By fine-tuning with BCM, our model can better estimate the motion in intermediate frames, thereby generating more reliable intermediate frames and achieving state-of-the-art performance.

\begin{figure*}[t]
  \centering
  \includegraphics[width=1\textwidth]{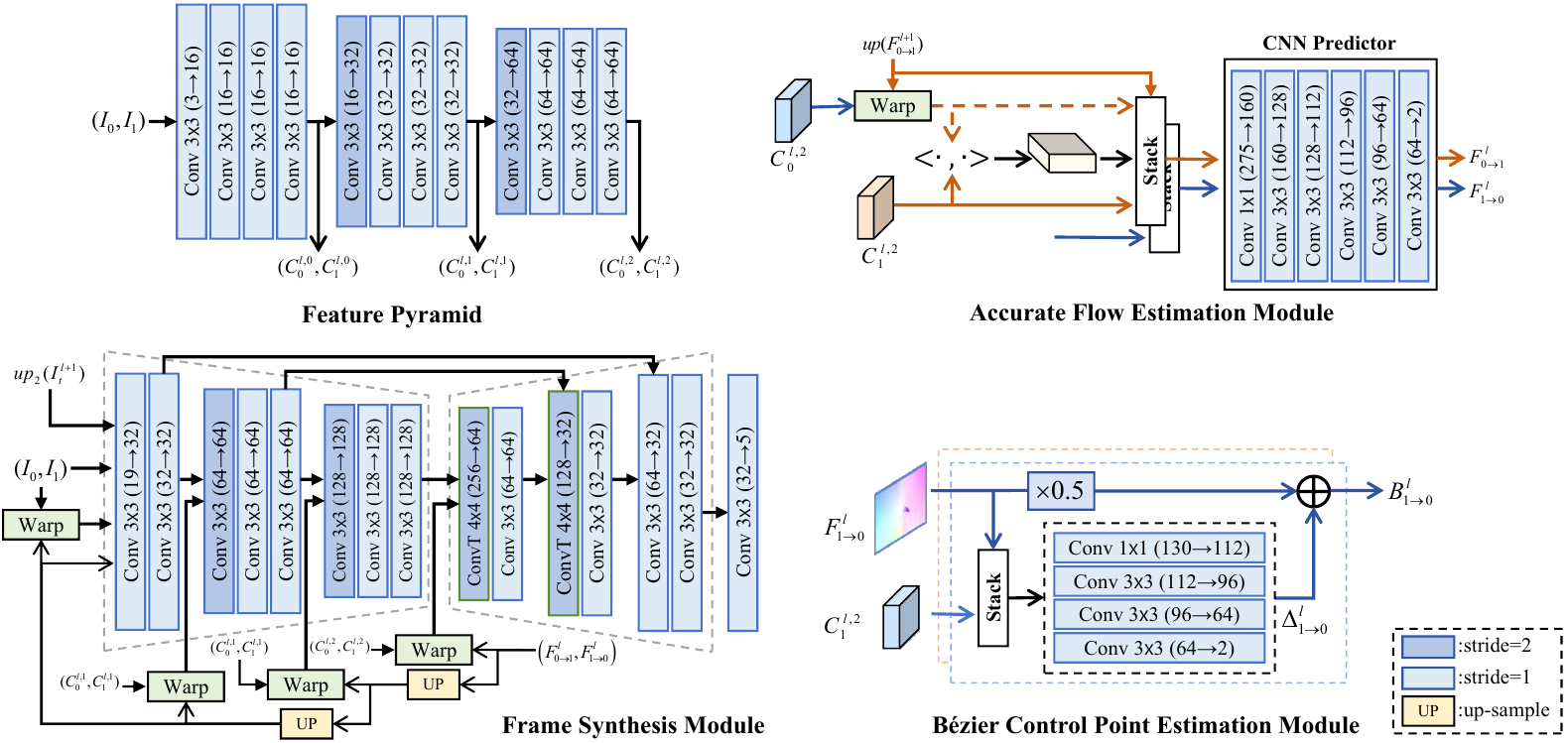}
  \caption{Architecture details of ABC-base.}
  \label{fig:arch}
\end{figure*}

\section{Experiments}
\begin{table*}[t]
  \centering
    \caption{Quantitative comparison for arbitrary-time interpolation.  [T] indicates that the model is trained end-to-end using the Vimeo90K-septuplet dataset. [D] refers to the use of the distance index proposed in \cite{zhong2023clearer} to eliminate speed ambiguity. [B] denotes our proposed method of using Bézier control points calculated by GT in the first stage of training to eliminate both speed and direction ambiguities. The best result is shown in \textcolor[rgb]{ 1,  0,  0}{\textbf{red}}, and the second best is \textcolor[rgb]{ .282,  .455,  .796}{\textbf{\underline{blue}}}.}
\begin{tabular}{lcccccccccc}
\toprule
\multirow{2}[4]{*}{Methods} & \multicolumn{2}{c}{Vimeo90K-septuplet} & \multicolumn{4}{c}{X-Test (2K)} & \multicolumn{4}{c}{X-Test (4K)} \\
\cmidrule(r){2-3} \cmidrule(r){4-7} \cmidrule(r){8-11}      & LPIPS$\downarrow$ & NIQE$\downarrow$ & PSNR$\uparrow$ & SSIM$\uparrow$ & LPIPS$\downarrow$ & NIQE$\downarrow$ & PSNR$\uparrow$ & SSIM$\uparrow$ & LPIPS$\downarrow$ & NIQE$\downarrow$ \\
\midrule
\text{[T]}RIFE~\cite{huang2022real} & 0.105  & \textcolor[rgb]{.282,.455,.796}{\textbf{\underline{6.663}}}  & 31.36  & 0.910  & 0.128  & 5.357  & 30.42  & 0.902  & 0.156  & 7.362  \\
\text{[T]}AMT-S~\cite{li2023amt} & 0.101  & 6.866  & 31.19  & 0.908  & 0.122  & \textcolor[rgb]{.282,.455,.796}{\textbf{\underline{5.443}}} & 30.30 & 0.902  & 0.149  & \textcolor[rgb]{.282,.455,.796}{\textbf{\underline{7.187}}} \\
\text{[T]}EMA-VFI~\cite{zhang2023extracting} & \textcolor[rgb]{.282,.455,.796}{\textbf{\underline{0.086}}}  & 6.736  & \textcolor[rgb]{.282,.455,.796}{\textbf{\underline{32.52}}}& \textcolor[rgb]{.282,.455,.796}{\textbf{\underline{0.929}}}& \textcolor[rgb]{.282,.455,.796}{\textbf{\underline{0.093}}}  & 5.180  & \textcolor[rgb]{.282,.455,.796}{\textbf{\underline{31.27}}} & 0.915  & \textcolor[rgb]{.282,.455,.796}{\textbf{\underline{0.147}}}  & 7.397 \\
\text{[T]}UPR-base~\cite{jin2023unified} & 0.095 & 6.751 & 31.83 & 0.921 & 0.108 & 5.328 & 30.83 & \textcolor[rgb]{.282,.455,.796}{\textbf{\underline{0.911}}} & 0.156 & 7.298 \\
\textbf{[T]ABC-base} & \textcolor[rgb]{1,0,0}{\textbf{0.073}}& \textcolor[rgb]{1,0,0}{\textbf{6.392}}& \textcolor[rgb]{1,0,0}{\textbf{32.73}}& \textcolor[rgb]{1,0,0}{\textbf{0.931}}& \textcolor[rgb]{1,0,0}{\textbf{0.069}}& \textcolor[rgb]{1,0,0}{\textbf{4.929}}& \textcolor[rgb]{1,0,0}{\textbf{31.43}} & \textcolor[rgb]{1,0,0}{\textbf{0.918}} & \textcolor[rgb]{1,0,0}{\textbf{0.100}}& \textcolor[rgb]{1,0,0}{\textbf{6.651}}\\
\midrule
\text{[D]}RIFE~\cite{zhong2023clearer} & 0.092  & 6.344& 31.72  & 0.917  & 0.090  & 4.972  & 30.44  & 0.904  & 0.129  & 7.097\\
\text{[D]}AMT-S~\cite{zhong2023clearer} & 0.090  & 6.452  & 30.40  & 0.907  & 0.101  & 5.127  & 29.61  & 0.901  & 0.142  & 7.152  \\
\text{[D]}EMA-VFI~\cite{zhong2023clearer} & 0.079 & 6.457  & \textcolor[rgb]{.282,.455,.796}{\textbf{\underline{32.40}}}& \textcolor[rgb]{1,0,0}{\textbf{0.931}}& 0.075 & 4.895 & \textcolor[rgb]{.282,.455,.796}{\textbf{\underline{31.30}}}  & \textcolor[rgb]{1,0,0}{\textbf{0.921}} & 0.130  & 7.148  \\
\text{[D]}ABC-base & 0.072 & 6.139  & 31.80 & 0.923 & 0.058 & 4.693 & 30.69  & 0.910 & 0.089  & 6.511  \\
BiM-VFI~\cite{seo2024bim}& \textcolor[rgb]{.282,.455,.796}{\textbf{\underline{0.070}}}& \textcolor[rgb]{.282,.455,.796}{\textbf{\underline{6.009}}}& 31.24& 0.918& \textcolor[rgb]{.282,.455,.796}{\textbf{\underline{0.044}}}& \textcolor[rgb]{1,0,0}{\textbf{4.387}}& 30.80& 0.913& \textcolor[rgb]{.282,.455,.796}{\textbf{\underline{0.068}}}&\textcolor[rgb]{.282,.455,.796}{\textbf{\underline{6.449}}}\\
\textbf{[B]UPR-base~\cite{jin2023unified}} & 0.077&  6.335& 31.74 & 0.920  & 0.065& 4.750& 30.88 & 0.911  & 0.098& 6.802\\
\textbf{[B]ABC-base} & \textcolor[rgb]{1,0,0}{\textbf{0.066}}& \textcolor[rgb]{1,0,0}{\textbf{5.990}}& \textcolor[rgb]{1,0,0}{\textbf{32.64}}& \textcolor[rgb]{.282,.455,.796}{\textbf{\underline{0.929}}}& \textcolor[rgb]{1,0,0}{\textbf{0.043}}& \textcolor[rgb]{.282,.455,.796}{\textbf{\underline{4.500}}}& \textcolor[rgb]{1,0,0}{\textbf{31.67}}& \textcolor[rgb]{.282,.455,.796}{\textbf{\underline{0.919}}}& \textcolor[rgb]{1,0,0}{\textbf{0.065}}& \textcolor[rgb]{1,0,0}{\textbf{6.372}}\\
\textit{ours([B] vs [T])} &
\textcolor[rgb]{1,0,0}{\textit{$\downarrow$9.6\%}} & \textcolor[rgb]{1,0,0}{\textit{$\downarrow$6.3\%}} & \textit{$\downarrow$0.3\%} & \textit{$\downarrow$0.2\%} & \textcolor[rgb]{1,0,0}{\textit{$\downarrow$37.7\%}} & \textcolor[rgb]{1,0,0}{\textit{$\downarrow$8.7\%}} & \textcolor[rgb]{1,0,0}{\textit{$\uparrow$0.8\%}} & \textcolor[rgb]{1,0,0}{\textit{$\uparrow$0.1\%}} & \textcolor[rgb]{1,0,0}{\textit{$\downarrow$35.0\%}} & \textcolor[rgb]{1,0,0}{\textit{$\downarrow$4.2\%}} \\
\bottomrule
\end{tabular}
  \label{tab:clear}%
\end{table*}%

\begin{figure*}[t]
    \centering
    \includegraphics[width=\textwidth]{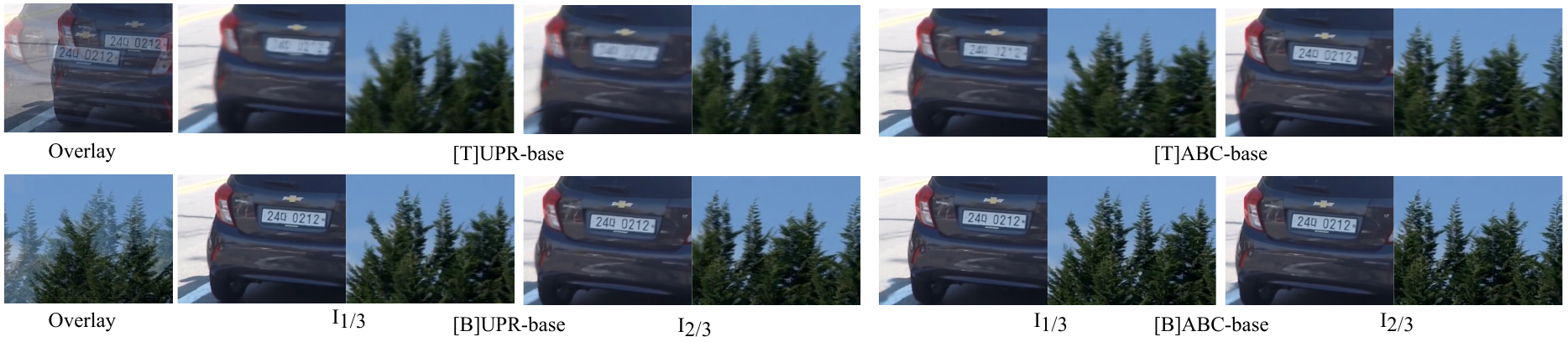}
  \caption{Qualitative comparison of different training settings between [T] and [B]. [B] shows clearer results.}
  \label{fig:TvB}
\end{figure*}

\begin{figure*}[t]
    \centering
    \includegraphics[width=\textwidth]{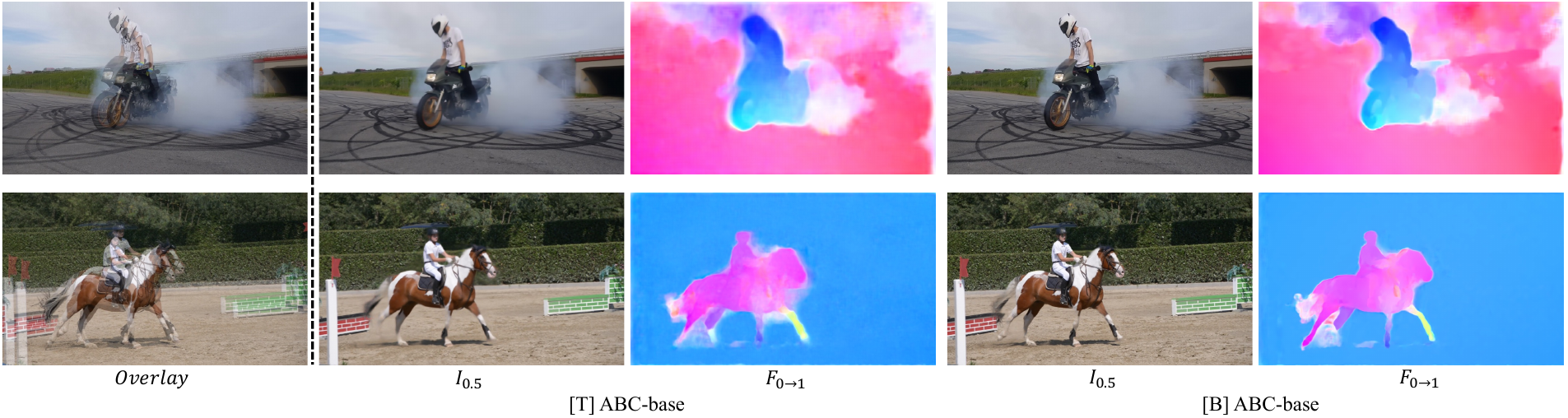}
  \caption{Visualization comparison of interpolation results and optical flow between different training settings ([T] and [B]) using ABC-base. [B] exhibits clearer results.}
  \label{fig:TvB_flow}
\end{figure*}

\begin{figure*}[t]
    \centering
    \includegraphics[width=\textwidth]{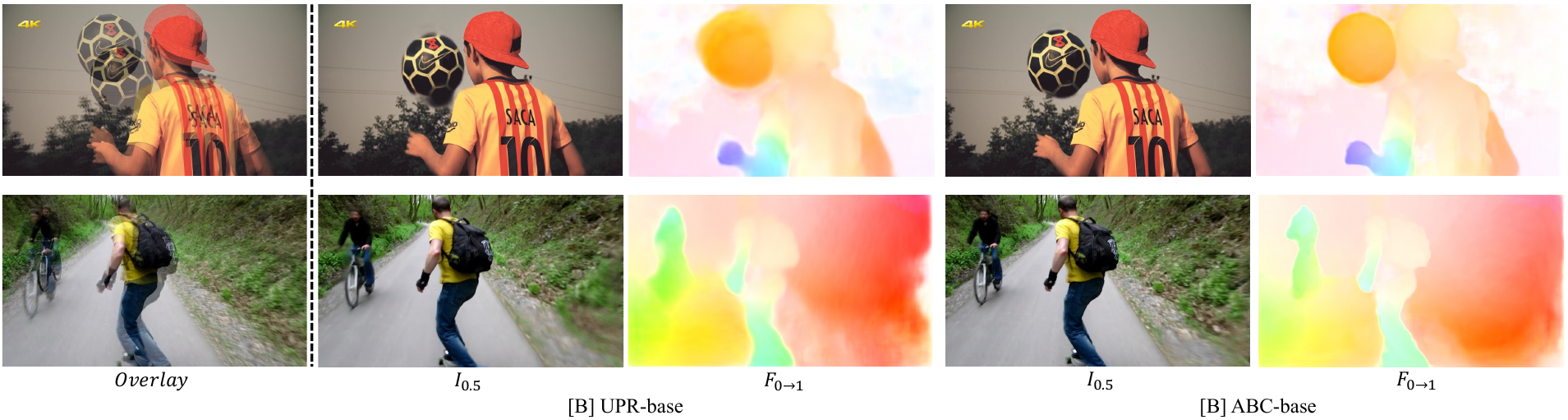}
  \caption{Visualization comparison of interpolation results and optical flow between different methods (UPR-base and ABC-base) under the same training setting [B]. ABC-base exhibits better results.}
  \label{fig:BvB_flow}
\end{figure*}

\begin{figure*}[t]
    \centering
    \includegraphics[width=\textwidth]{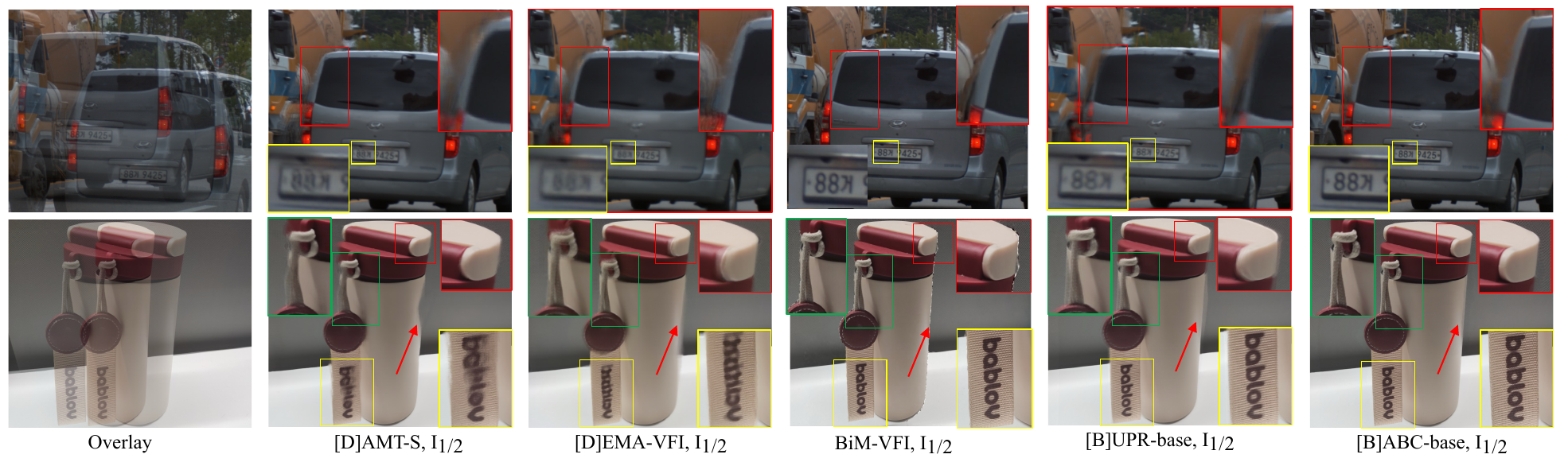}

  \caption{Qualitative comparison of different training settings between [D]~\cite{zhong2023clearer}, BiM-VFI~\cite{seo2024bim} and [B]. [B] shows clearer results.}
  \label{fig:DvB}
\end{figure*}

\begin{table*}[t]
  \centering
  \caption{Quantitative comparison (PSNR/SSIM) with SOTA methods.
  We measure the average latency over 1000 iterations at 1280×720 resolution using a single NVIDIA RTX 3090. 
  Meanwhile, we classify the models into two categories based on whether their inference latency exceeds 100 ms/f, and conduct separate comparisons for each category.
  }
\resizebox{0.9\linewidth}{!}{
\begin{tabular}{lccccccc}
\toprule
\multirow{2}[4]{*}{Method} & \multirow{2}[4]{*}{Vimeo90K-triplet} & \multicolumn{2}{c}{SNU-FILM} &\multicolumn{2}{c}{Xiph} & Latency 
& Params  \\
\cmidrule(r){3-4} \cmidrule(r){5-6}     &       & Medium & Hard  & 2K    & 4K     & (ms/f) & (M)  \\
\midrule
ToFlow~\cite{xue2019video}  & 33.73/0.968   & 34.39/0.974   & 28.44/0.918   & 33.93/0.922& 30.74/0.856  & 88
& 1.4 \\
M2M-VFI~\cite{hu2022many}  & 35.49/0.978   & 35.74/0.980   & 30.32/0.936   & 36.44/\textcolor[rgb]{ .282,  .455,  .796}{\textbf{\underline{0.943}}} & 33.92/0.899  & 40
& 7.6 \\
RIFE~\cite{huang2022real}  & 35.65/0.978   & 35.75/0.979   & 30.10/0.933   & 36.19/0.938 & 33.76/0.894  & 29
& 9.8 \\
IFRNet-B~\cite{kong2022ifrnet} & 35.80/0.979 & 35.94/0.979   & 30.41/0.936   & 36.00/0.936 & 33.99/0.893  & 30
& 5 \\
AMT-S~\cite{li2023amt} & 35.97/\textcolor[rgb]{ 1,  0,  0}{\textbf{{0.983}}}  & 35.98/\textcolor[rgb]{ 1,  0,  0}{\textbf{{0.983}}}  & 30.60/\textcolor[rgb]{ 1,  0,  0}{\textbf{{0.940}}}  & 36.11/0.940 & 34.29/0.901  & 51
& 3 \\
EMA-S~\cite{zhang2023extracting} & \textcolor[rgb]{ .282,  .455,  .796}{\textbf{\underline{36.07}}}/0.980 & 35.88/0.980 & \textcolor[rgb]{ .282,  .455,  .796}{\textbf{\underline{30.69}}}/0.938 & 36.55/0.942 & 34.25/0.902 & 76 & 14.5\\
UPR-base~\cite{jin2023unified}  & 36.03/0.980 & \textcolor[rgb]{ .282,  .455,  .796}{\textbf{\underline{36.16}}}/0.980   & 30.67/0.937   &    \textcolor[rgb]{ .282,  .455,  .796}{\textbf{\underline{36.60}}}/0.943   &   \textcolor[rgb]{ .282,  .455,  .796}{\textbf{\underline{34.30}}}/\textcolor[rgb]{ .282,  .455,  .796}{\textbf{\underline{0.903}}}     & 65
& 1.7 \\
ABC-base  & \textcolor[rgb]{ 1,  0,  0}{\textbf{{36.29}}}/\textcolor[rgb]{ .282,  .455,  .796}{\textbf{\underline{0.981}}}  & \textcolor[rgb]{ 1,  0,  0}{\textbf{{36.26}}}/\textcolor[rgb]{ .282,  .455,  .796}{\textbf{\underline{0.980}}}  & \textcolor[rgb]{ 1,  0,  0}{\textbf{{30.87}}}/\textcolor[rgb]{ .282,  .455,  .796}{\textbf{\underline{0.938}}}  & \textcolor[rgb]{ 1,  0,  0}{\textbf{{36.84}}}/\textcolor[rgb]{ 1,  0,  0}{\textbf{{0.944}}} & \textcolor[rgb]{ 1,  0,  0}{\textbf{{34.55}}}/\textcolor[rgb]{ 1,  0,  0}{\textbf{{0.904}}}  & 81
& 1.8 \\
\midrule
VFIFormer~\cite{lu2022video}& 36.50/0.982   & 36.09/0.980   & 30.67/0.938   & OOM   & OOM    & 1293
& 24.1 \\
TTVFI~\cite{liu2023ttvfi}      & 33.44/0.970 & 34.06/0.972 & 29.92/0.934 & 34.78/0.962 & 33.32/0.944 & 1517& 16.6 \\
EMA-VFI~\cite{zhang2023extracting}${^{\dag}}$  & 36.50/0.980   & 35.86/0.979   & 30.80/0.938   & 36.74/0.944 & 34.55/\textcolor[rgb]{ .282,  .455,  .796}{\textbf{\underline{0.906}}}  & 211
& 66 \\
AMT-G~\cite{li2023amt}  & 36.53/0.982& 36.12/\textcolor[rgb]{ 1,  0,  0}{\textbf{{0.981}}}  & 30.78/\textcolor[rgb]{ 1,  0,  0}{\textbf{{0.939}}}  & 36.38/0.941 & 34.63/0.904  & 250
& 30.6 \\
PerVFI~\cite{wu2024perception} & 33.97/0.970 & 34.62/0.974 & 29.79/0.928 & 34.59/0.916 & 32.24/0.867    & 608
& 13.9 \\
SGM-1/2-points~\cite{liu2024sparse} & 35.82/0.979 & 36.20/0.980 & 30.76/0.937 &  36.48/0.942 & 34.17/0.905  & 387
& 20.9 \\
  VTinker~\cite{wu2026vtinker}    & 35.19/0.976 & 35.48/0.977 & 30.29/0.933 & 35.88/0.959 & 33.49/0.933 & 184 & 60.6 \\
  EDEN~\cite{zhang2025eden}        & 32.67/0.959 & 34.70/0.974 & 29.62/0.928 & 33.35/0.946 & 31.40/0.919 & 243 & 157.9 \\
  LDMVFI~\cite{danier2024ldmvfi}     & 33.09/0.963 & 34.03/0.971 & 28.56/0.918 & 33.87/0.949 & 31.43/0.921 & 7044 & 439.0 \\
  TLB-VFI~\cite{lyu2025tlb}     & 32.78/0.960 & 34.28/0.971 & 29.47/0.926 & 33.71/0.946 & 31.68/0.919 & 595 & 46.7 \\
  UPR-LARGE~\cite{jin2023unified}  & 36.42/0.982   & \textcolor[rgb]{ .282,  .455,  .796}{\textbf{\underline{36.29}}}/0.980 & \textcolor[rgb]{ .282,  .455,  .796}{\textbf{\underline{30.86}}}/0.938 &  \textcolor[rgb]{ .282,  .455,  .796}{\textbf{\underline{37.07}}}/\textcolor[rgb]{ .282,  .455,  .796}{\textbf{\underline{0.945}}}     &  \textcolor[rgb]{ .282,  .455,  .796}{\textbf{\underline{34.65}}}/0.905      & 147
& 6.6 \\
   IQ-VFI~\cite{hu2024iq}& \textcolor[rgb]{ 1,  0,  0}{\textbf{{36.60}}}/0.982& 36.24/0.980& 30.83/0.938& 36.68/0.942& 34.72/0.905& -&-\\
ABC-LARGE  & \textcolor[rgb]{ .282,  .455,  .796}{\textbf{\underline{36.59}}}/\textcolor[rgb]{ 1,  0,  0}{\textbf{{0.982}}}& \textcolor[rgb]{ 1,  0,  0}{\textbf{{36.36}}}/\textcolor[rgb]{ .282,  .455,  .796}{\textbf{\underline{0.980}}}  & \textcolor[rgb]{ 1,  0,  0}{\textbf{{30.96}}}/\textcolor[rgb]{ .282,  .455,  .796}{\textbf{\underline{0.938}}}  & \textcolor[rgb]{ 1,  0,  0}{\textbf{{37.12}}}/\textcolor[rgb]{ 1,  0,  0}{\textbf{{0.946}}} & \textcolor[rgb]{ 1,  0,  0}{\textbf{{34.72}}}/\textcolor[rgb]{ 1,  0,  0}{\textbf{{0.906}}}  & 181& 7.2 \\
\bottomrule
\end{tabular}%

}

  \label{tab:Sota}%
\end{table*}%

\begin{table}[t]
  \centering
  \caption{Quantitative comparison on multiple metrics under SNU-FILM Medium and Hard subsets.}
  \resizebox{\linewidth}{!}{
  \begin{tabular}{lcccccc}
  \toprule
  \multirow{2}{*}{Method} 
  & \multicolumn{3}{c}{SNU-FILM Medium} 
  & \multicolumn{3}{c}{SNU-FILM Hard} \\
  \cmidrule(r){2-4} \cmidrule(r){5-7}
  & PSNR$\uparrow$ & SSIM$\uparrow$ & tOF$\downarrow$ 
  & PSNR$\uparrow$ & SSIM$\uparrow$ & tOF$\downarrow$ \\
  \midrule
  AMT-G~\cite{li2023amt} & 36.12 & \textcolor[rgb]{ .282,  .455,  .796}{\textbf{\underline{0.981}}} & 0.677 & 30.78 & \textcolor[rgb]{ 1,  0,  0}{\textbf{{0.939}}} & 1.955 \\
  EMA-VFI~\cite{zhang2023extracting} & 35.86 & 0.979 & 0.539 & 30.80 & \textcolor[rgb]{ .282,  .455,  .796}{\textbf{\underline{0.938}}} & 1.756 \\
  UPR-LARGE~\cite{jin2023unified} & \textcolor[rgb]{ .282,  .455,  .796}{\textbf{\underline{36.29}}} & 0.980 & \textcolor[rgb]{ .282,  .455,  .796}{\textbf{\underline{0.531}}} & \textcolor[rgb]{ .282,  .455,  .796}{\textbf{\underline{30.86}}} & \textcolor[rgb]{ .282,  .455,  .796}{\textbf{\underline{0.938}}} & \textcolor[rgb]{ .282,  .455,  .796}{\textbf{\underline{1.627}}} \\
  VTinker~\cite{wu2026vtinker} & 35.48 & 0.977 & 0.540 & 30.29 & 0.933 & 1.637 \\
  TTVFI~\cite{liu2023ttvfi} & 34.06 & 0.972 & 0.671 & 29.92 & 0.934 & 1.805 \\
  EDEN~\cite{zhang2025eden} & 34.70 & 0.974 & 0.554 & 29.62 & 0.928 & 1.729 \\
  LDMVFI~\cite{danier2024ldmvfi} & 34.03 & 0.971 & 0.623 & 28.56 & 0.918 & 2.160 \\
  TLB-VFI~\cite{lyu2025tlb} & 34.28 & 0.971 & 0.768 & 29.47 & 0.926 & 2.379 \\
  ABC-LARGE & \textcolor[rgb]{ 1,  0,  0}{\textbf{{36.59}}} & \textcolor[rgb]{ 1,  0,  0}{\textbf{{0.982}}} & \textcolor[rgb]{ 1,  0,  0}{\textbf{{0.529}}} & \textcolor[rgb]{ 1,  0,  0}{\textbf{{30.96}}} & \textcolor[rgb]{ .282,  .455,  .796}{\textbf{\underline{0.938}}} & \textcolor[rgb]{ 1,  0,  0}{\textbf{{1.579}}} \\
  \bottomrule
  \end{tabular}
  }
  \label{tab:temporal_snu}
\end{table}

\begin{table}[htbp]
  \centering
  \caption{Quantitative comparisons on X-Test \cite{sim2021xvfi} for 4$\times$  interpolation. Here, $I_0$, $I_{16}$, and $I_{32}$ in X-Test correspond to the input frames $I_0$, $I_{0.5}$, and $I_1$, respectively. \text{[M]} indicates that we use multi-frame to estimate flow and obtain the Bézier control points.}
  \resizebox{\linewidth}{!}{
    \begin{tabular}{lccccc}
    \toprule
    \multirow{2}[1]{*}{Methods} &\multirow{2}[1]{*}{Input Frames}& \multicolumn{2}{c}{X-Test (2K)}& \multicolumn{2}{c}{X-Test (4K)}\\
\cmidrule(lr){3-4}  \cmidrule(lr){5-6}   && PSNR& SSIM& PSNR & SSIM\\
    \midrule

     \text{[T]}ABC-base&$(I_{0/1},I_{0.5})$& 36.18&0.962&34.89& 0.952 \\
    \text{[T,M]}ABC-base&$(I_0, I_{0.5}, I_1)$& 36.38&0.961&34.99& 0.951\\
    \text{[B]}ABC-base&$(I_{0/1},I_{0.5})$& 36.36&0.965&34.98& 0.954 \\
    \text{[B,M]}ABC-base&$(I_0, I_{0.5}, I_1)$& \textcolor[rgb]{ 1,  0,  0}{\textbf{37.07}}&\textcolor[rgb]{ 1,  0,  0}{\textbf{0.967}}&\textcolor[rgb]{ 1,  0,  0}{\textbf{35.55}}& \textcolor[rgb]{ 1,  0,  0}{\textbf{0.956}}\\
    \bottomrule
    \end{tabular}%
    }
  \label{tab:multi}%
\end{table}%

\begin{figure}[htbp]
    \centering
    \includegraphics[width=\linewidth]{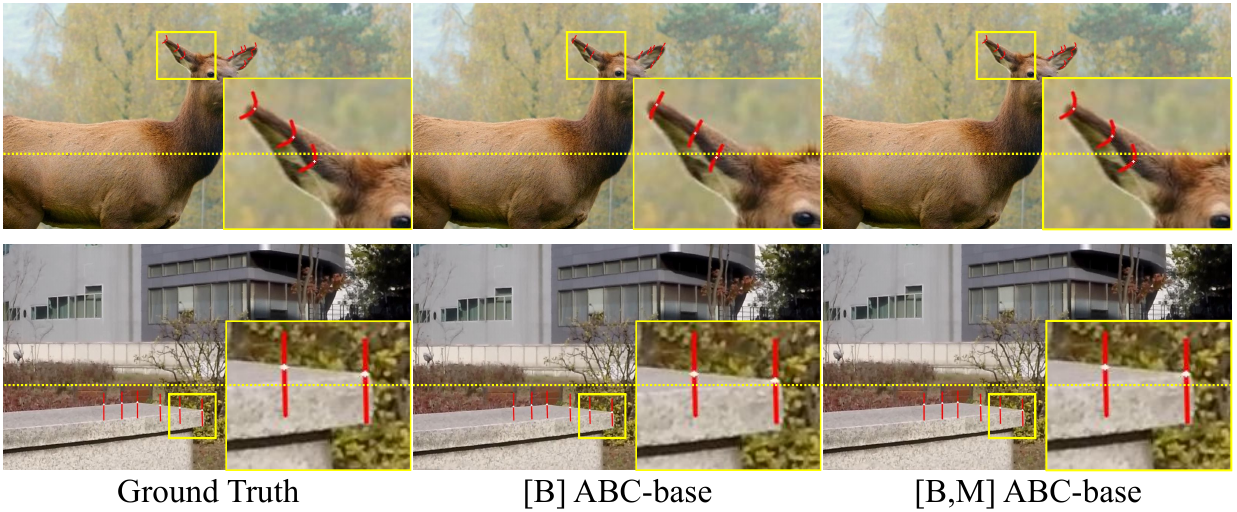}
    \caption{Qualitative comparison of different inference settings. In \text{[M]}, using an additional frame to calculate Bézier control points achieves better consistency with GT motion in direction and speed. \textcolor[rgb]{.7, .54, 0}{\textbf{The yellow dotted line}} denotes the reference line, and \textcolor{red}{\textbf{the red line}} denotes the trajectory of the key point.}
    \label{fig:uniform}
\end{figure}

We conducted the first and second stages of training using the Vimeo90K-septuplet~\cite{xue2019video} and Vimeo90K-triplet~\cite{xue2019video} datasets, respectively.
For each stage, the model is trained for 800K iterative steps using the AdamW optimizer~\cite{loshchilov2017decoupled} on 4 NVIDIA RTX V100 GPUs.
Following the training settings of UPR-Net~\cite{jin2023unified}, the total batch size is 32, and the model’s learning rate decay follows the cosine attenuation schedule from ${2 \times {10^{ - 4}}}$ to ${2 \times {10^{ - 5}}}$. 
Similarly, we follow the augmentation pipeline including random cropping, reversing, flipping, rotating, pyramid levels, and randomly cropping patches with size ${256 \times 256}$ in UPR-Net~\cite{jin2023unified}. Additionally, we use the loss of UPR-Net~\cite{jin2023unified} to train our ABC-Inter, which is the sum of Charbonnier loss $\rho$~\cite{charbonnier1994two} and census loss $L_{cen}$~\cite{meister2018unflow} between ground truth ${I_t^{GT}}$ and the interpolation ${{I_t}}$ estimated at the bottom pyramid level:
\begin{equation}
L = \rho (I_t^{GT} - {I_t}) + {L_{cen}}(I_t^{GT},{I_t}),
\label{eq:loss}
\end{equation}

\subsection{Benchmarks}
Following a common strategy, ABC-Inter is trained only on Vimeo90K~\cite{xue2019video} datasets, and evaluated on a variety of benchmarks at different resolutions.
We use pixel-centric metrics (PSNR and SSIM~\cite{wang2004image}), perceptual metrics (LPIPS~\cite{zhang2018unreasonable} and NIQE~\cite{mittal2012making}), and the temporal consistency metric tOF~\cite{chu2020learning} for evaluation.
\begin{itemize}
    \item \textbf{Vimeo90K~\cite{xue2019video}}: Vimeo90K is the most commonly used training or evaluation benchmark in VFI. The Vimeo dataset includes two parts: triplet and septuplet, with a resolution of ${448 \times 256}$.

    \item \textbf{X4K1000FPS~\cite{sim2021xvfi}}: X4K1000FPS is a 4K resolution benchmark with a wide range of motions, and was clipped as X-Test for a metric, which enables multi-frame (${\times 8}$) interpolation. 

    \item \textbf{SNU-FILM~\cite{choi2020channel}}: SNU-FILM dataset contains 1,240 triplets at different resolution (around ${1280 \times 720}$). It contains four subsets with different levels of motion scales-easy, medium, hard, and extreme.

    \item  \textbf{Xiph~\cite{xphi1994}}: Xiph contains eight video sequences with 4K resolution. Following~\cite{niklaus2020softmax}, we downsample and center-crop the original images to 2K resolution to obtain the ``Xiph-2K'' and ``Xiph-4K''.
\end{itemize}

\subsection{Model Variants}
To facilitate better application and comparison, we construct two models of different sizes: ABC-base and ABC-LARGE. Taking ABC-base as an example, its specific model structure is detailed in Fig.\ref{fig:arch}. The three stages of the feature encoder have channel numbers of 16, 32, and 64 respectively. The six-layer channels of the flow estimation module are 160, 128, 112, 96, 64, 2. The four-layer channels of the Bézier control point estimation module are 112, 96, 64, 2. The channel numbers of the three encoder stages of synthesis module are 32,64,128. Compared to ABC-base, the channel numbers of each module in ABC-LARGE are multiplied by 2. The number of convolutional layers in each module has not changed.

\subsection{Comparison with the SOTAs}
In order to verify the effectiveness of our proposed method, we evaluate the models trained in our two stages separately.
\textbf{Training stage one.}
At this stage, we design an AFM module and introduce the use of GT-computed Bézier control points in training to eliminate motion ambiguity within the training dataset. This enables our model to produce clear results and estimate more accurate optical flow for the frame interpolation. 
To better demonstrate the advantages of our method, we first report the performance metrics of existing state-of-the-art (SOTA) models trained end-to-end on Vimeo90K-septuplet for arbitrary frame interpolation in Tab.~\ref{tab:clear}, showing their results on Vimeo90K-septuplet and X-Test. This setting is denoted as [T]
Subsequently, we presented the results of InterpAny, BiM-VFI, and ABC-Inter in resolving motion ambiguity within the training set using different methods. Among these, InterpAny employed distance indices, denoted as [D], while ABC-Inter utilized Bézier control points, denoted as [B].

Under the training setting of [T], our ABC-base model achieves the best performance in terms of pixel-centric metrics, with a 0.48 dB improvement in PSNR over the state-of-the-art (SOTA) models on X-Test (4K). This demonstrates the effectiveness of our AFM module compared to UPR-Net~\cite{jin2023unified}.

Compared to the setting of [D] or BiM-VFI, our proposed [B] training method achieves significant improvements in perceptual metrics on X-Test. 
To better verify that the introduction of Bézier control points enables us to generate clearer results, we do not make any modifications to the UPR-Net architecture. Instead, we only introduce Bézier control points during the training process. As can be seen from Tab.~\ref{tab:clear}, compared with [T] UPR-base, [B] UPR-base achieves significant improvements in perceptual metrics across different datasets. This verifies the effectiveness of our proposed Bézier control points.

We provide visual comparisons to analyze the effect of introducing B\'ezier control points. As shown in Fig.~\ref{fig:TvB} and Fig.~\ref{fig:TvB_flow}, the [B] setting produces clearer interpolation results and more accurate optical flow than the [T] setting for both UPR-Net and ABC-Inter. Moreover, Fig.~\ref{fig:BvB_flow} shows that ABC-base further improves the interpolation quality and flow accuracy over UPR-base under the same [B] setting, validating the effectiveness of AFM. We also compare [B] with distance indexing [D] and BiM-VFI, where ABC-Inter achieves better results in large-motion scenes.

The above method assuming uniform motion during inference may not match the GT motion. Thus, we can introduce an additional frame to calculate Bézier control points for modeling intermediate frame motion. As shown in Tab~\ref{tab:multi}, using multiple frames enables [B,M]ABC-base to achieve a 0.71dB PSNR gain over [B]ABC-base on X-Test (2K). However, the improvement under the [T] setting is smaller, due to less accurate flow estimation caused by unremoved motion blur in the training set.
Meanwhile, in Fig.~\ref{fig:uniform}, we present the visualization results using multiple frames and the motion trajectories of key points. It can be seen that using multiple frames enables the interpolation of non-linear motion between two frames during inference.

\begin{figure*}[h]
    \centering
    \includegraphics[width=\linewidth]{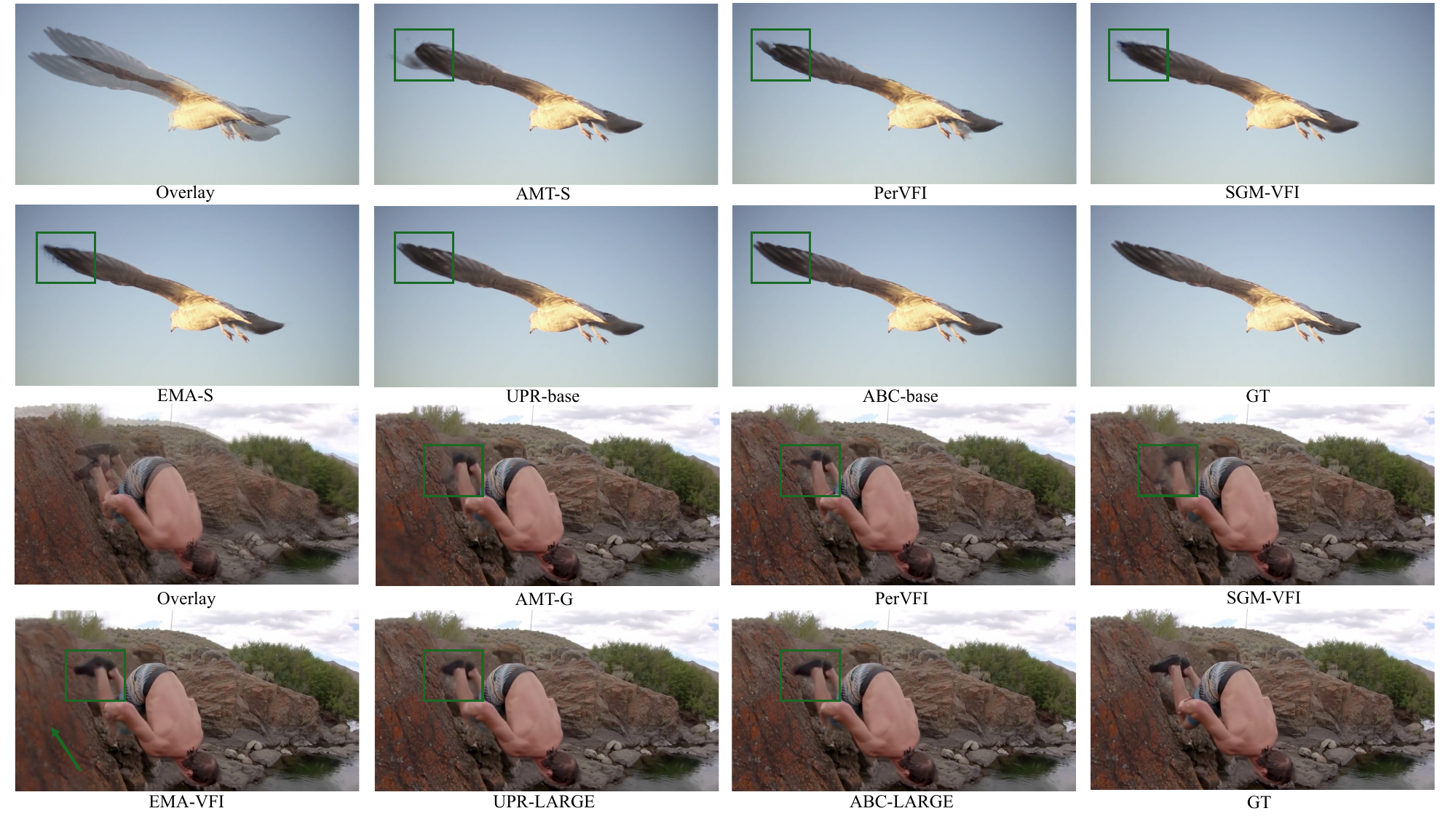}
    \caption{Visual comparisons between ABC-Inter and other methods on SNU-FILM dataset}
    \label{fig:vis_snu}
\end{figure*}

\begin{figure*}[h]
    \centering
    \includegraphics[width=\linewidth]{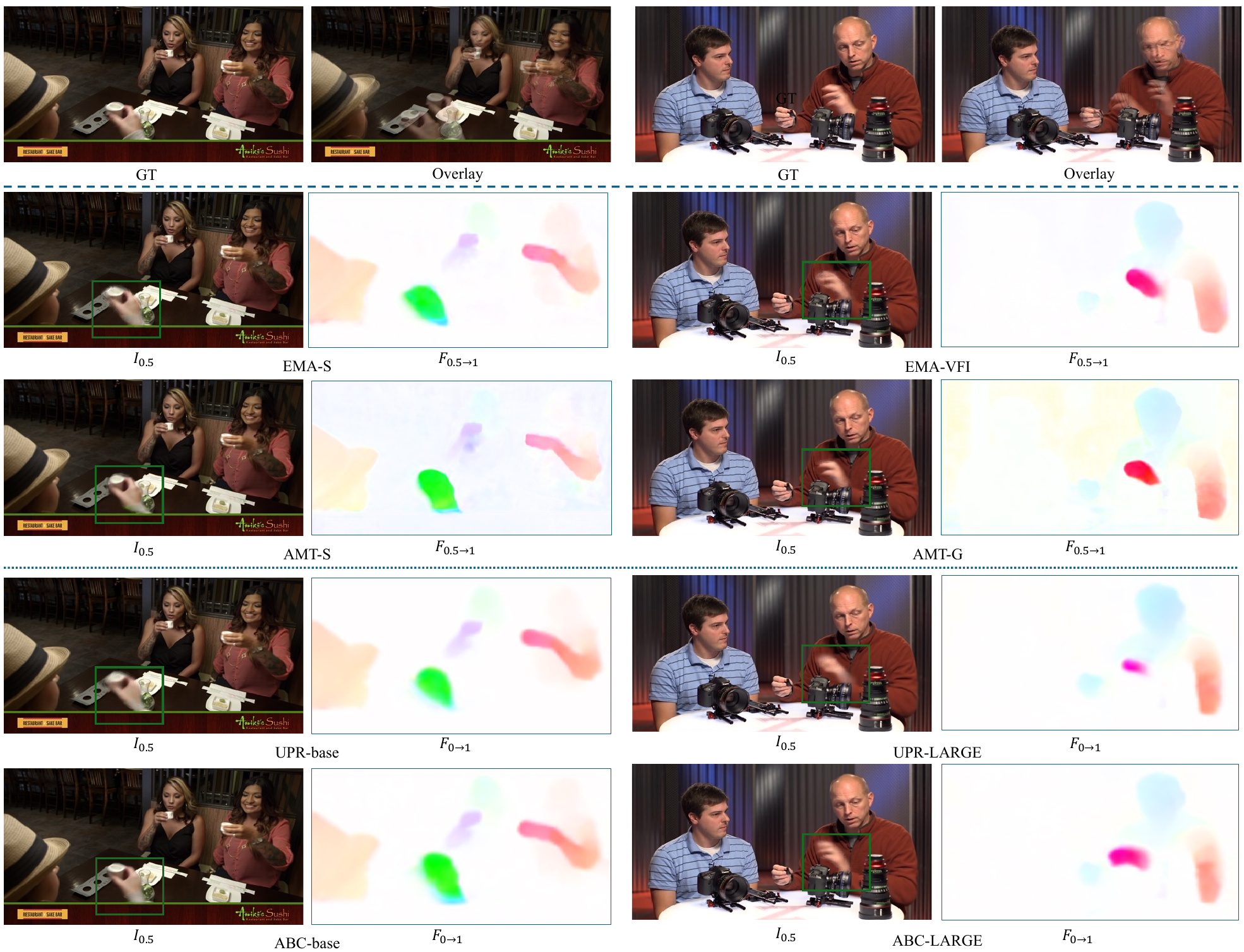}
    \caption{Visual comparisons between ABC-Inter and other methods on Vimeo-90K dataset}
    \label{fig:vis_vimeo}
\end{figure*}

\textbf{Training stage two.}
In this stage, we utilize the model trained in the first stage as the initial model, introduce the BCM module to estimate the Bézier control points, and then retrain the model on the Vimeo90K-triplet.
In Tab.~\ref{tab:Sota}, we classify models based on whether their inference latency exceeds 100 ms/frame to show efficient frame interpolation performance.
Table~\ref{tab:temporal_snu} compares ABC-LARGE with recent VFI methods on SNU-FILM Medium and Hard. ABC-LARGE achieves the best PSNR and tOF, showing strong reconstruction quality and temporal consistency.

\begin{figure*}[h]
    \centering
    \includegraphics[width=\linewidth]{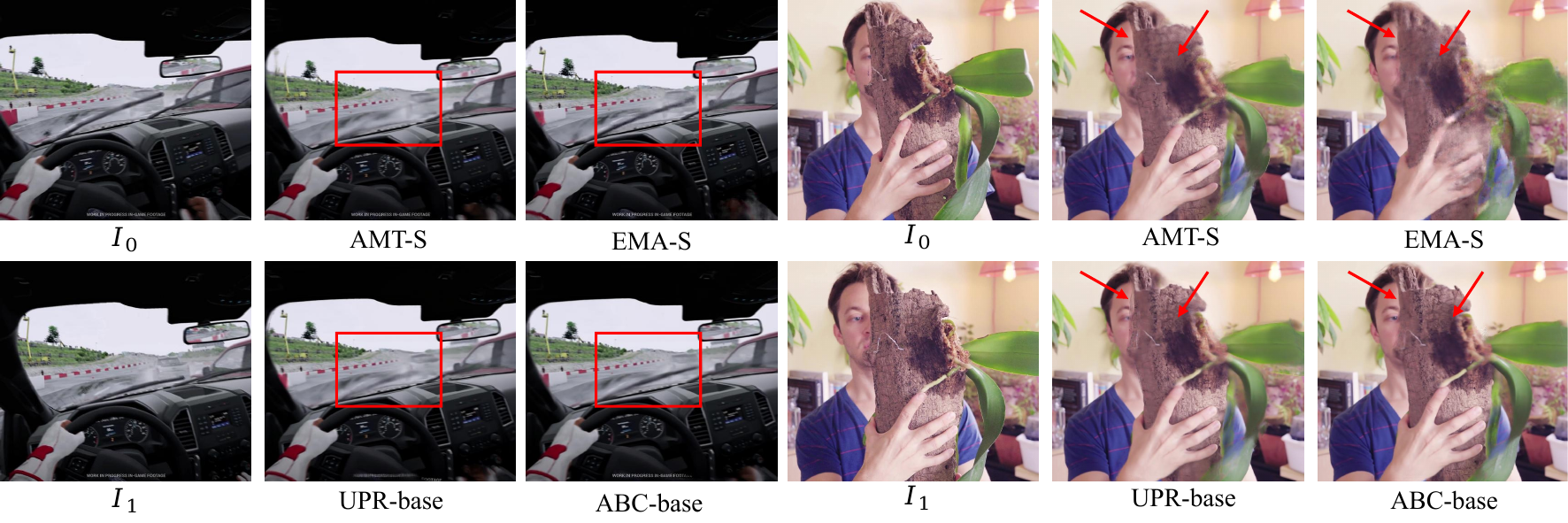}
    \caption{Visual comparison under challenging scenarios, including motion blur, occlusion, and extremely large motion. Although these cases are difficult for current VFI methods, ABC-Inter produces clearer structures and more stable interpolation results than other compared methods.}
    \label{fig:extreme_case}
\end{figure*}

In addition, we provide visualizations for different methods. In Fig~\ref{fig:vis_snu}, we present the visualization results of ABC-base and ABC-LARGE on the SNU-FILM dataset, from which it can be seen that the results generated by our methods are clearer.
We also show the visualized intermediate frames and optical flows of different methods on the Vimeo-90K dataset in Fig~\ref{fig:vis_vimeo}. Among these methods, EMA-VFI and AMT adopt the backward warping approach, so $F_{t->1}$ is displayed; while UPR-Net and ABC-Inter use the forward warping approach, thus $F_{0->1}$ is presented. It can be observed that the flow estimated by ABC-Inter is more accurate and clean, and at the same time, it can generate more accurate intermediate results.
In addition, we provide visual comparisons under more challenging scenarios. As shown in Fig.~\ref{fig:extreme_case}, ABC-Inter may still suffer from slight blur or structural artifacts under severe motion blur, occlusion, and extremely large motion. Nevertheless, compared with AMT-S, EMA-S, and UPR-base, our method produces clearer structures and more stable interpolation results.

\begin{table}
  \centering
 \caption{Ablation (PSNR/SSIM) of our designs. We validate the effectiveness of two-stage training strategy with BCM and AFM.}  
  \resizebox{\linewidth}{!}{
    \begin{tabular}{llc}
    \toprule
    Experiments & Case & Vimeo \\
    \midrule
    \multirow{2}[2]{*}{Training \& BCM} & only first stage & 35.26/0.978 \\
        & w/o first stage & 36.18/0.981 \\
        & two stage & \textbf{36.29/0.981} \\
    \midrule
    \multirow{6}[6]{*}{AFM} & w/o correlation volume & 36.05/0.980 \\
        & w/ correlation volume & \textbf{36.18/0.981} \\
\cmidrule{2-3}        & forward warping  & 35.87/0.979 \\
        & backward warping  & \textbf{36.18/0.981} \\
\cmidrule{2-3}        & mid-frame alignment & 36.11/0.980  \\
        & input-frames alignment & \textbf{36.18/0.981} \\
    \midrule
    \multirow{2}[2]{*}{BCM} & w/o BCM & 36.10/0.980  \\
        & w/ BCM (second-order) & 36.18/0.981 \\
        & w/ BCM (third-order) & \textbf{36.19/0.981} \\
    \bottomrule
    \end{tabular}%
    }
  \label{tab:ablationstudy}
\end{table}%

\begin{figure}[t]
    \centering
    \includegraphics[width=\linewidth]{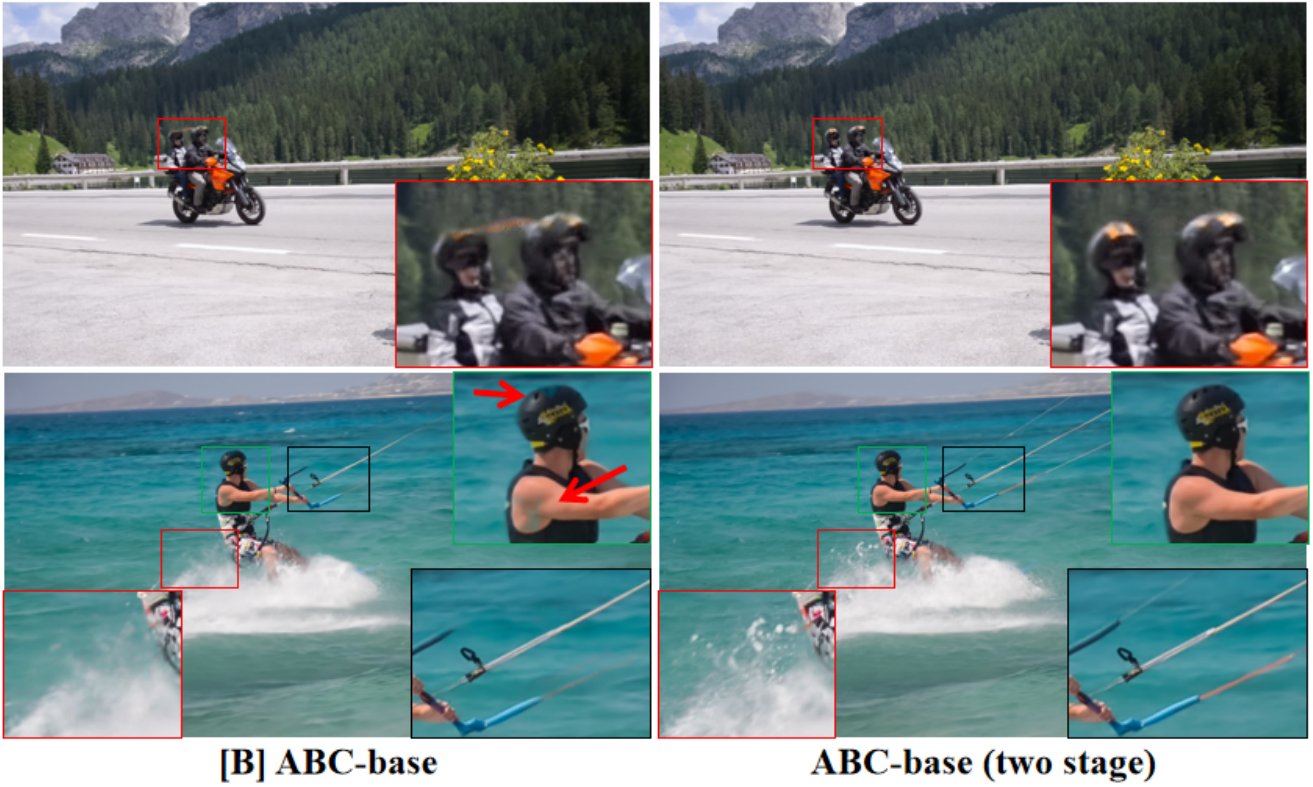}
    \caption{Visualization comparison between the first-stage [B]ABC-base and the second-stage ABC-base.}
    \label{fig:1v2}
\end{figure}
\begin{figure}[t]
    \centering
    \includegraphics[width=\linewidth]{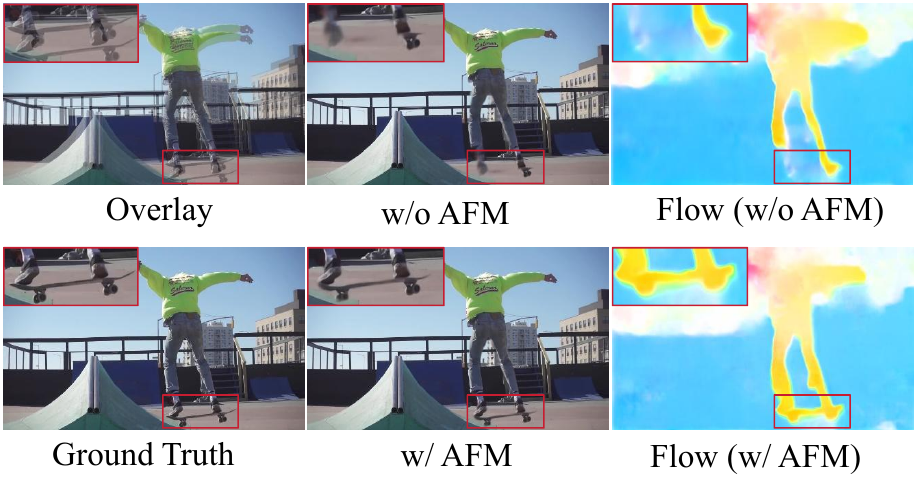}
    \caption{Visualization of AFM ablation. The w/o AFM version uses the flow estimation module from UPR-Net.}
    \label{fig:afm}
\end{figure}

\begin{figure}[t]
  \centering
  \includegraphics[width=\linewidth]{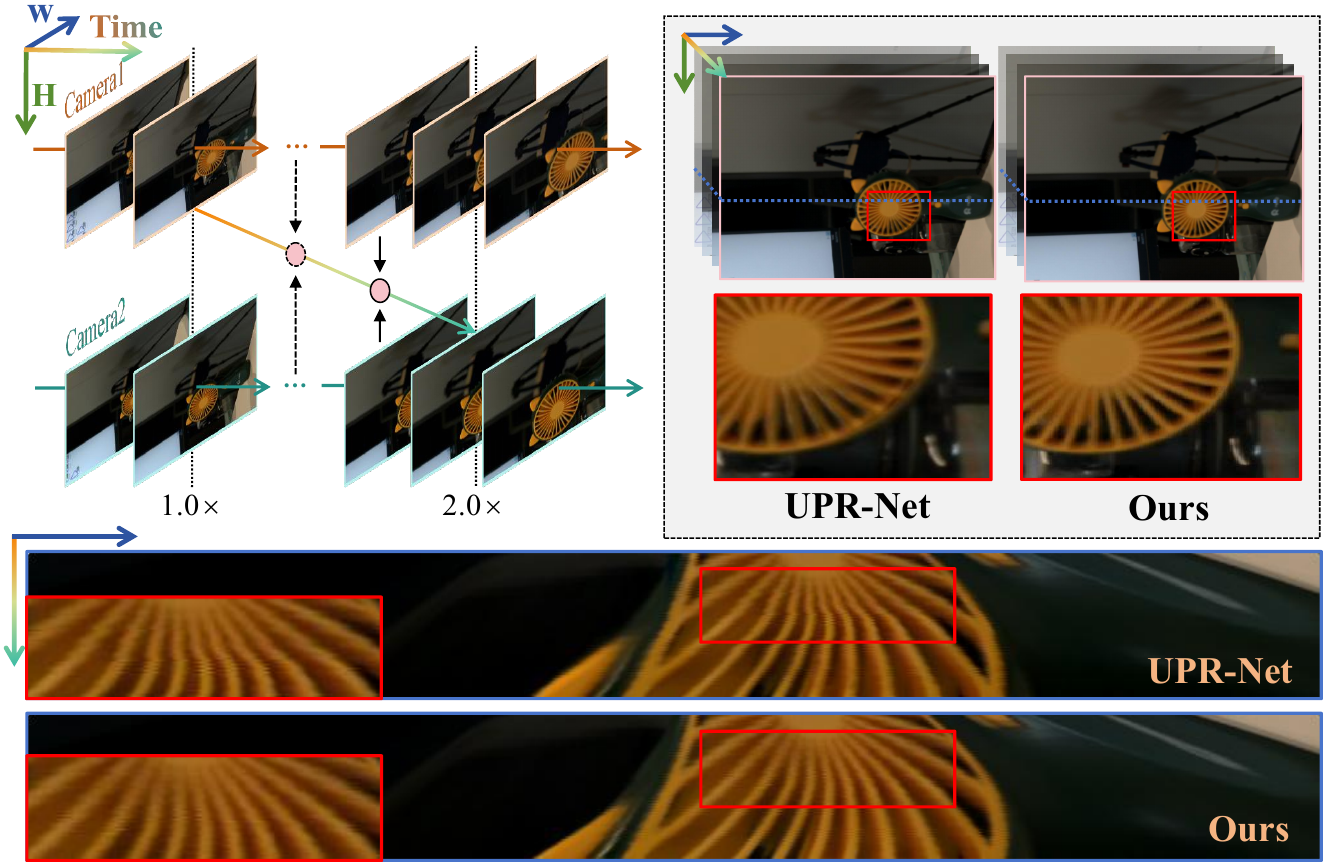}
  \caption{Application in SAT. Two streams are captured by the super-wide and wide lens 1${\&}$2 of smartphone OPPO Find X6, with smooth transitions through UPR-Net~\cite{jin2023unified} and our ABC-Inter. Compared to the UPR-Net~\cite{jin2023unified}, Our ABC-Inter has a much higher image quality and video smoothness.}
  \label{fig:oppo}
\end{figure}

\subsection{Ablation Study}

In this section, we conduct ablation experiments to validate the effectiveness of ABC-Inter. 
All ablation experiments are performed on ABC-base and validated on Vimeo90K~\cite{xue2019video}.
The specific ablation results can be found in Tab.~\ref{tab:ablationstudy}. 

\textbf{Training Strategy \& BCM:} 
In Tab.~\ref{tab:clear} and Fig.~\ref{fig:TvB}, \ref{fig:TvB_flow}, \ref{fig:BvB_flow}, we validate the effectiveness of using B\'ezier control points in the first stage to reduce motion ambiguity. We further evaluate the BCM module and the two-stage training strategy in Tab.~\ref{tab:ablationstudy}. The results show that using only the first-stage model suffers from the linear motion assumption during inference, while directly training BCM without the first stage is affected by motion ambiguity. In contrast, the complete two-stage strategy first learns accurate task-oriented flows and then estimates B\'ezier control points with BCM, achieving the best performance. The visualization in Fig.~\ref{fig:1v2} further shows that BCM improves inter-frame motion modeling and produces better interpolation results.

\textbf{Flow Estimation Module:}
The qualitative comparison with UPR-Net~\cite{jin2023unified} in Tab.~\ref{tab:clear} has preliminarily demonstrated the effectiveness of the AFM module. Furthermore, we demonstrate in Fig.~\ref{fig:afm_comp} that using AFM yields more accurate optical flow compared with UPR-Net. To further verify the effectiveness of the specific design of AFM, we modify the relevant structures and conduct single-stage training on Vimeo-triplet, as detailed below.
1) \textbf{Correlation Volume (CV):} Removing the CV component from the flow estimation module results in a performance decline across all benchmark tests, consistent with the findings in ~\cite{sun2018pwc}. This suggests that the design of CV can better leverage local information to update the flow. 2) \textbf{Warping Method:} Changing the method of aligning coordinates in the module to forward warping significantly impacts the model's performance. This can be attributed to the inconsistency between the flow used for warping and the flow intended for updating. 3) \textbf{Aligning Method: } we attempt flow estimation for aligning coordinates in intermediate frames like UPR-Net~\cite{jin2023unified}. However, it does not yield satisfactory performance, likely due to the misalignment between flow coordinates and intermediate frame coordinates. As shown in Fig.~\ref{fig:afm}, using AFM can produce more accurate flow and generate better results.

\textbf{B´ezier Control Point Estimation:} We ablate the order of the Bézier curve used in BCM. The results show that the second-order Bézier curve provides a consistent performance improvement, improving the result from 36.10/0.980 to 36.18/0.981. Using a third-order Bézier curve only brings a marginal gain, reaching 36.19/0.981. This suggests that higher-order Bézier curves do not provide obvious additional benefits, and the second-order formulation offers a good balance between performance and simplicity.

\subsection{Applications}

Our method performs well across multiple applications, such as animation frame interpolation, slow-motion generation and smart transitions (SAT). Fig.~\ref{fig:oppo} presents the results of our method in SAT. 
When shooting video with a multi-camera phone, zoom functionality is achieved through switching between lenses of different focal lengths.  However, due to the parallax effect between lenses, the transitions may appear unnatural.
The existing algorithm achieves transition by directly switching or blending between the two camera frames, which seriously affects the user experience.
To enhance the natural zoom experience, VFI algorithms are employed for multi-camera SAT.
Based on the multi-camera function of OPPO Find X6, we apply ABC-Inter to SAT, by interpolating frames from multi-cameras to achieve smooth transitions during video recording. As shown in Fig.~\ref{fig:oppo}, compared to UPR-Net~\cite{jin2023unified}, our ABC-Inter can get better zoom quality and smoothness to enhance the user experience in a greater way, which shows its promising potential in generating continuous video sequences.

\section{Conclusion}

To address the issues of motion ambiguity in training sets and the inaccuracies caused by uniform motion assumptions during inference, this work proposes ABC-Inter, an accurate motion estimation algorithm utilizing Bézier control points. 
ABC-Inter introduces an \textbf{Accurate Flow estimation Module (AFM)}, which decouples bidirectional flow estimation, ensuring coordinate consistency for more precise optical flow.
Furthermore, ABC-Inter explicitly models non-linear motion by introducing \textbf{Bézier control points}.
Specifically, it utilizes ground-truth intermediate frames during the first training stage to calculate Bézier control points, effectively eliminating motion ambiguity. 
For inference, the model adopts a flexible strategy. 
It can either \textbf{introduce additional input frames} to directly calculate control points \textbf{without retraining} or employ a dedicated \textbf{Bézier Control point estimation Module (BCM)} fine-tuned in a second stage to predict non-linear motion from two frames.
Due to its excellent motion modeling capabilities, ABC-Inter achieves SOTA on multiple benchmarks and exhibits excellent visual perception.
\bibliographystyle{plain}
\bibliography{ref}

\vspace{-2em}

\begin{IEEEbiography}[{\includegraphics[width=1in,height=1.25in,clip,keepaspectratio]{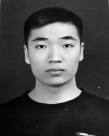}}]{Shuhao Han}
received the B.S. degree from Hefei University of Technology, China, in 2023. He is currently pursuing the Ph.D. degree with the Media Computing Laboratory, Nankai University, supervised by Prof. Chun-Le Guo. His research interests include image super-resolution, video frame interpolation and image quality assessment.
\end{IEEEbiography}
\vspace{-2em}

\begin{IEEEbiography}[{\includegraphics[width=1in,height=1.25in,clip,keepaspectratio]{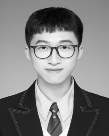}}]{Chenyang Wu}
 received the B.S. degree from Guizhou University, China, in 2024. He is currently pursuing the Ph.D. degree with the Media Computing Laboratory, Nankai University, supervised by Prof. Chongyi Li. His research interests include video frame interpolation and video completion.
\end{IEEEbiography}
\vspace{-2em}

\begin{IEEEbiography}[{\includegraphics[width=1in,height=1.25in,clip,keepaspectratio]{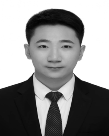}}]{Chun-Le Guo}
 received the Ph.D. degree from Tianjin University, China, under the
supervision of Prof. Ji-Chang Guo. He was a
Visiting Ph.D. Student with the School of Elec
tronic Engineering and Computer Science, Queen
Mary University of London (QMUL), U.K. He
was a Research Associate with the Department of
Computer Science, City University of Hong Kong
(CityUHK). He was a Postdoctoral Researcher with
Nankai University, under the guidance of Prof.
Ming-Ming Cheng. He is currently an Associate Pro
fessor with Nankai University. His research interests include image processing,
computer vision, and deep learning.
\end{IEEEbiography}
\vspace{-2em}

\begin{IEEEbiography}[{\includegraphics[width=1in,height=1.25in,clip,keepaspectratio]{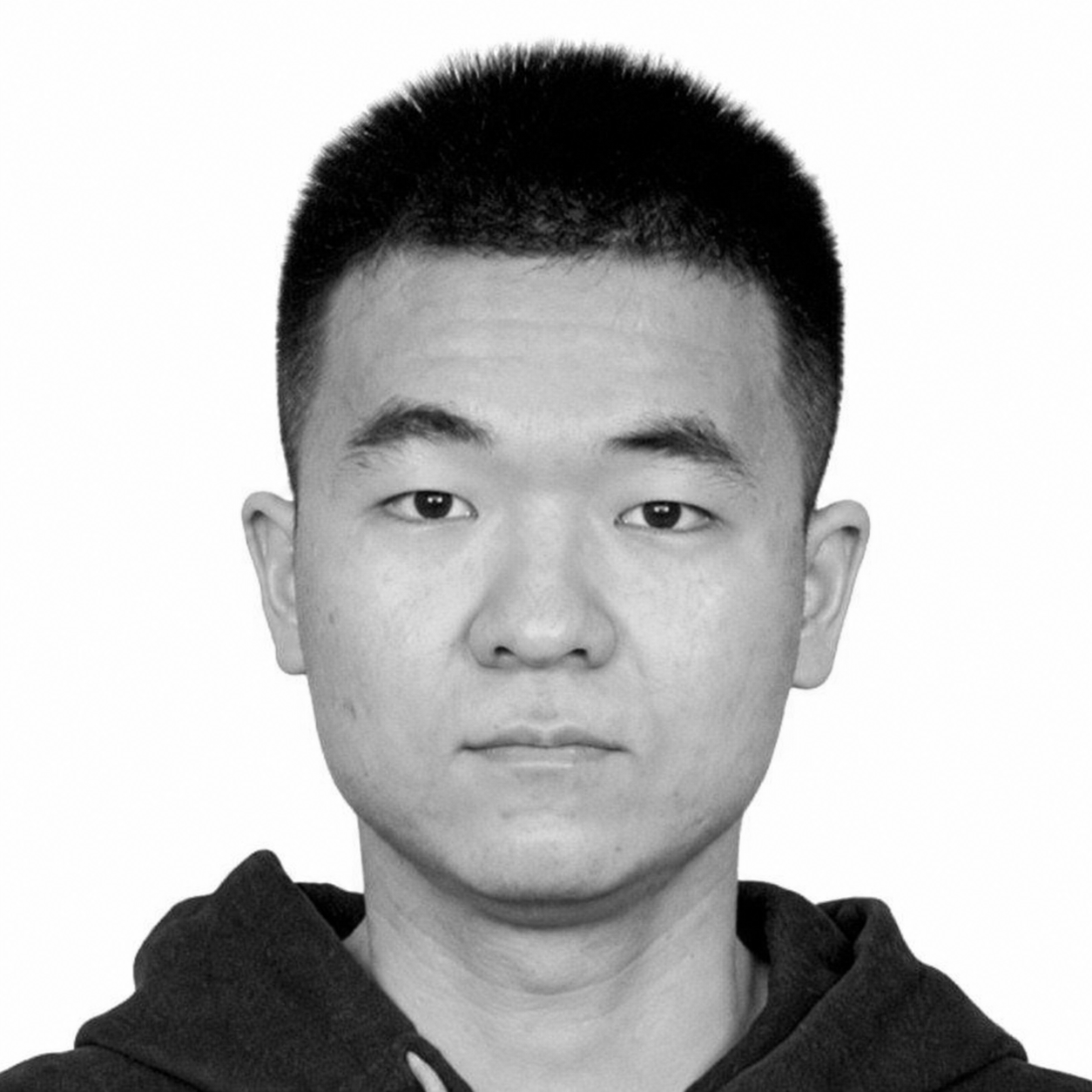}}]{Zheng-Peng Duan}
received the B.S. degree from Xidian University, China, in 2022. He is currently pursuing the Ph.D. degree with the Media Computing Laboratory, Nankai University, supervised by Prof. Chongyi Li and Prof. Ming-Ming Cheng. His research interests include image restoration and interactive editing.
\end{IEEEbiography}
\vspace{-2em}

\begin{IEEEbiography}[{\includegraphics[width=1in,height=1.25in,clip,keepaspectratio]{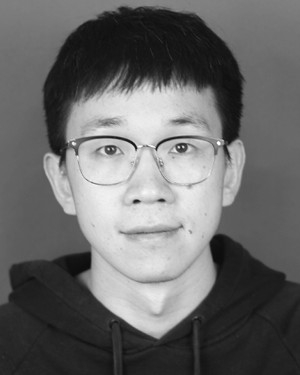}}]{Zhen Li}
received his PhD degree from the College of Computer Science, Nankai University, under the joint supervision of Prof. Ming-Ming Cheng and Prof. Xiu-Li Shao. He currently works at Alibaba Inc. His research interests include computer vision and deep learning, particularly focusing on image/video restoration and enhancement, generation and editing, etc.
\end{IEEEbiography}
\vspace{-2em}

\begin{IEEEbiography}[{\includegraphics[width=1in,height=1.25in,clip,keepaspectratio]{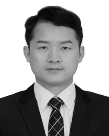}}]{Ming-Ming Cheng}
 received his PhD degree
from Tsinghua University in 2012. Then he did 2 years research fellow, with Prof. Philip Torr in Oxford. He is now a professor at Nankai University, leading the Media Computing Lab. His research interests include computer graphics, computer vision, and image processing. He received research awards including National Science Fundfor Distinguished Young Scholars and
ACM China Rising Star Award. He is on the editorial boards of IEEE TPAMI and IEEE TIP.
\end{IEEEbiography}
\vspace{-2em}

\begin{IEEEbiography}[{\includegraphics[width=1in,height=1.25in,clip,keepaspectratio]{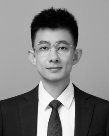}}]{Chongyi Li}
was a Research Assistant Professor with Nanyang Technological University, Singapore, and a Research Fellow with Nanyang Technological University and the City University of Hong Kong. He is currently a Professor with Nankai University, Tianjin, China. His research interests include low-level vision and computational photography.
\end{IEEEbiography}

\vfill

\end{document}